\documentclass[letterpaper, 10 pt, conference]{ieeeconf}  % Comment this line out if you need a4paper

\IEEEoverridecommandlockouts                              % This command is only needed if 
\usepackage{hyperref}
\usepackage{array}
\usepackage{amssymb}
\usepackage{amsmath}
\usepackage{amsfonts}
\usepackage{booktabs}
\usepackage{colortbl}
\usepackage{graphicx}
\usepackage{verbatim}
\usepackage{wrapfig}
\usepackage{xspace}
\usepackage{multirow}
\usepackage[scaled=0.85]{beramono}
\usepackage[T1]{fontenc}
\newcolumntype{P}[1]{>{\centering\arraybackslash}p{#1}}
\usepackage{caption}
\usepackage{siunitx} 
\usepackage{color, soul} 
\usepackage{float}
\newcommand{\method}{GD2P\xspace}

\usepackage{array}
\usepackage{amssymb}
\usepackage{amsmath}
\usepackage{amsfonts}
\usepackage{booktabs}
\usepackage{colortbl}
\usepackage{graphicx}
\usepackage{verbatim}
\usepackage{wrapfig}
\usepackage{xspace}
\usepackage{multirow}
\usepackage[scaled=0.85]{beramono}
\usepackage[T1]{fontenc}
\newcolumntype{P}[1]{>{\centering\arraybackslash}p{#1}}
\usepackage{caption}
\usepackage{siunitx} 
\usepackage{color, soul} 
\usepackage{float}
\usepackage{stfloats}
\usepackage{cuted}  
\usepackage{amsmath}

\usepackage[backend=biber,style=ieee,maxbibnames=6,minbibnames=1]{biblatex}
\newcommand{\numtrials}{840\xspace}

\definecolor{lightgreen}{rgb}{0.8, 1.0, 0.8}
\definecolor{Reddish_Purple}{rgb}{0.800, 0.475, 0.655}
\definecolor{pinegreen}{RGB}{28, 210, 177}

\newcommand{\remark}[3]{{\color{#2}[#1: #3]}}
\newcommand{\daniel}[1]{\remark{Daniel}{cyan}{#1}}

\title{\LARGE \bf
Learning Geometry-Aware Nonprehensile Pushing and Pulling \\
with Dexterous Hands\\
}
\author{Yunshuang Li, Yiyang Ling, Gaurav S. Sukhatme, Daniel Seita % <-this % 
\thanks{All authors are with the Thomas Lord Department of Computer Science at
the University of Southern California, USA.}%
\thanks{
Correspondence: {yunshuan}@usc.edu}%
}

\begin{document}
\maketitle
\thispagestyle{empty}
\pagestyle{empty}

\begin{strip}
\vspace*{-1.6cm}
\centering
\includegraphics[width=1.0\textwidth]{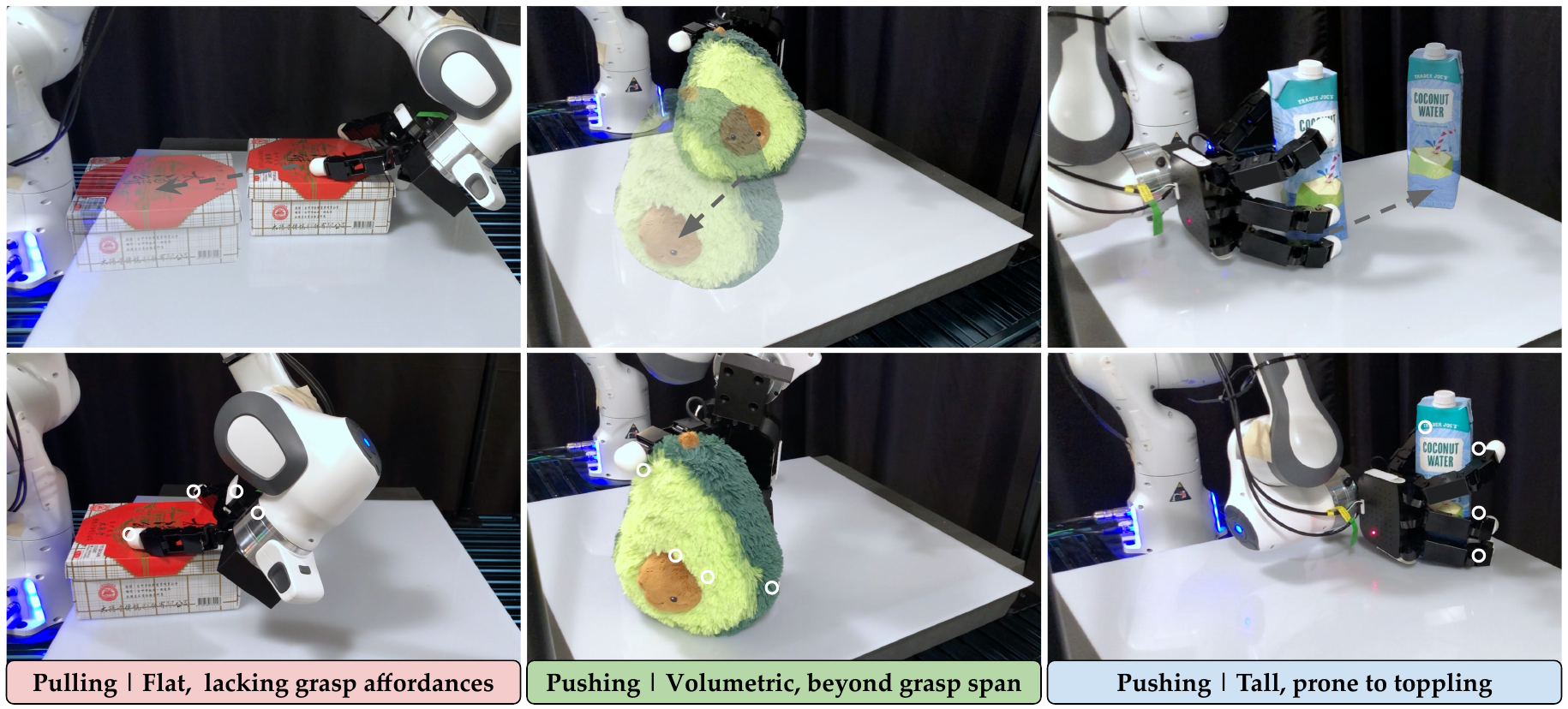} 
\captionof{figure}{
Three examples of nonprehensile manipulation using \method with a 4-finger, 16-DOF Allegro Hand. The top row shows the starting object configuration with its goal rendered as a transparent overlay, while the bottom row shows the result after the robot's motion. \method synthesizes diverse hand poses conditioned on object geometry, handling flat (left), volumetric (middle), and tall (right) objects. Grey arrows represent the transporting direction, whereas white volumetric dots mark the estimated fingertip contact with the object.}
\label{fig:pull}
\vspace*{-0.2cm}
\end{strip}

\begin{abstract}
    Nonprehensile manipulation, such as pushing and pulling, enables robots to move, align, or reposition objects that may be difficult to grasp due to their geometry, size, or relationship to the robot or the environment. Much of the existing work in nonprehensile manipulation relies on parallel-jaw grippers or tools such as rods and spatulas. In contrast, multi-fingered dexterous hands offer richer contact modes and versatility for handling diverse objects to provide stable support over the objects, which compensates for the difficulty of modeling the dynamics of nonprehensile manipulation.
       Therefore, we propose \underline{G}eometry-aware \underline{D}exterous \underline{P}ushing and \underline{P}ulling (\method) for nonprehensile manipulation with dexterous robotic hands. We study pushing and pulling by framing the problem as synthesizing and learning pre-contact dexterous hand poses that lead to effective manipulation. We generate diverse hand poses via contact-guided sampling, filter them using physics simulation, and train a diffusion model conditioned on object geometry to predict viable poses. 
    At test time, we sample hand poses and use standard motion planners to select and execute pushing and pulling actions. 
    We perform extensive real-world experiments with an Allegro Hand and a LEAP Hand, demonstrating that \method offers a scalable route for generating dexterous nonprehensile manipulation motions with its applicability to different hand morphologies.
    %We further demonstrate \method on a LEAP Hand, highlighting its applicability to different hand morphologies. 
    % Our pre-trained models and dataset, including 1.3 million hand poses across 2.3k objects, will be open-source to facilitate further research.  
    Our project website is available at: \href{https://geodex2p.github.io/}{\textit{geodex2p.github.io}}. 

\end{abstract}

% Two or three meaningful keywords should be added here
% \keywords{ Nonprehensile manipulation, dexterous hand} 

%===============================================================================

\section{Introduction}

\begin{figure*}[t]
\center
\includegraphics[width=1.0\textwidth]{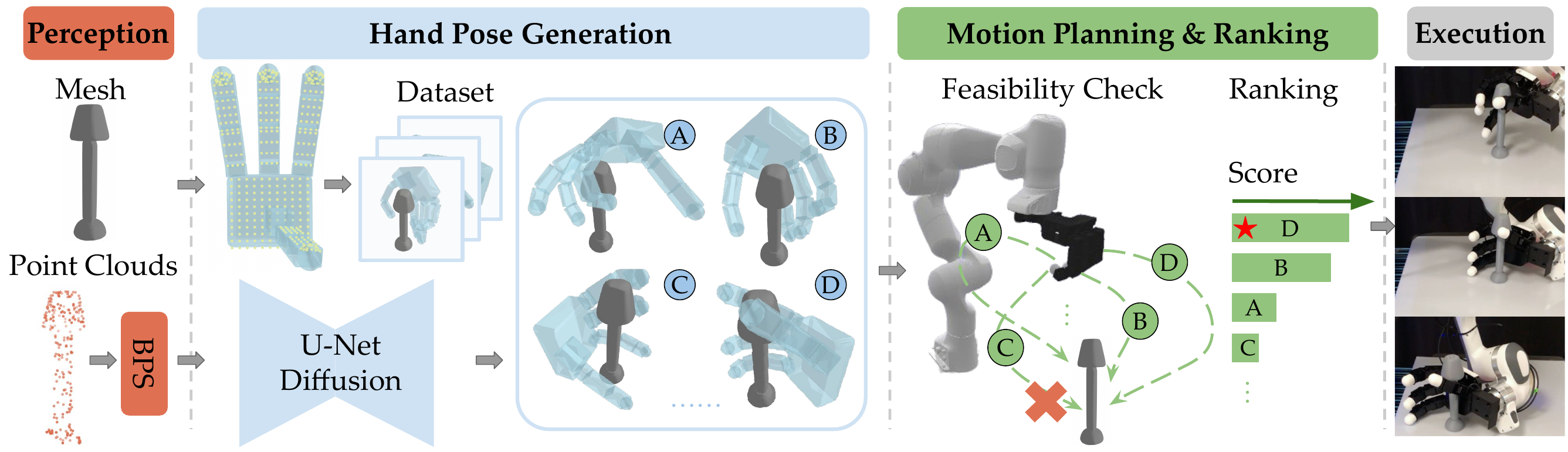} 
\caption{
Overview of \method. We present a large-scale dataset of hand poses specifically for pushing or pulling, and leverage it to train a diffusion model. During execution time, given an object, we obtain its basis point set representation~\cite{prokudin2019BPS} and pass that to our trained diffusion model, which uses the architecture from~\cite{weng2024dexdiffuser}. This model synthesizes diverse floating pre-contact hand poses formed from our large-scale data generation pipeline (Sec.~\ref{ssec:hand_poses}). Given these hand poses, we then check their feasibility in a physics simulator by adding the arm back in and performing motion planning~\cite{sundaralingam2023curobo}. We rank the feasible hand poses (e.g., ``C'' is infeasible in the example here since the motion planner detects an unavoidable collision between the arm and the table.) and select the best performing one (e.g., ``D'' in our example with no collision detected, while optimally facilitating the pushing direction.) and execute it in the real world. 
}
\label{fig:method}
\vspace*{-10pt}
\end{figure*}  

%Pushing and pulling objects are nonprehensile actions that are fundamental to how humans and robots interact with the physical world~\cite{mason1986nonprehensile,lynch1996nonprehensile,mason1999nonprehensile,lynchmason1999nonprehensile}. 
Nonprehensile actions are fundamental to how humans and robots interact with the physical world~\cite{mason1986nonprehensile,lynch1996nonprehensile,mason1999nonprehensile,lynchmason1999nonprehensile}. 
These actions permit the manipulation of objects that may be too large, heavy, or geometrically complex to grasp directly. 
While there has been tremendous progress in nonprehensile robot manipulation~\cite{zhou2022ungraspable,zhou2023hacman,jiang2024hacmanpp,cho2024corn,lyu2025dywa}, most work uses simple end-effectors such as parallel-jaw grippers, rods~\cite{zhang2024adaptigraph,chi2023diffusionpolicy}, or spatulas~\cite{wang2023dynamicresolution}. In contrast, multi-fingered hands with high degrees-of-freedom (DOF) such as the Allegro Hand or LEAP Hand~\cite{shaw2023leaphand} enable contact patterns that can be especially useful for stabilizing complex, awkward, or top-heavy objects, or for coordinating contact across multiple objects, compensating for the challenges of modeling nonprehensile manipulation dynamics. 
However, despite their promise and recent progress~\cite{wang2025dexterousnonprehensile}
% ~\cite{zhong2025dexgraspvlavisionlanguageactionframeworkgeneral}
, leveraging high-DOF hands for nonprehensile manipulation remains relatively underexplored due to the challenges of modeling hand-object relationships and planning feasible contact-rich motions. 
%\cite{wang2025dexterousnonprehensile}

In this paper, we study pushing and pulling objects using the 4-finger, 16-DOF Allegro and LEAP Hands. We select pushing and pulling as representative tasks of nonprehensile manipulation because they are more commonly used for manipulating general daily objects and are
more amenable to scaling. 
% , whereas motions like “rolling” or
% “tilting” may restrict the range of applicable objects. For
% example, rolling typically assumes cylindrical geometries.
Our insight is to recast this problem into one of synthesizing effective pre-contact hand poses, an approach inspired by recent success in generating large-scale datasets for dexterous manipulation~\cite{lum2024gripmultifingergraspevaluation,wang2023dexgraspnet,zhang2024dexgraspnet,xu2023unidexgrasp,wan2023unidexgrasp++,li2024multigrasp}.
We propose a scalable pipeline for generating hand poses for pushing and pulling objects. This involves contact-guided optimization and validation via GPU-accelerated physics simulation with IsaacGym~\cite{makoviychuk2021isaac}.
These filtered hand poses are then used to train a generative diffusion model conditioned on object geometry, represented using basis point sets~\cite{prokudin2019BPS}. 

At test time, we use visual data to reconstruct an object mesh in physics simulation. The trained diffusion model uses this mesh to generate diverse hand poses for pushing or pulling. We then validate the resulting hand poses in simulation, and execute the best-performing action in the real world. 
We call this pipeline \textbf{G}eometry-aware \textbf{D}exterous \textbf{P}ushing and \textbf{P}ulling (\method) with multi-fingered hands. 
Fig.~\ref{fig:pull} shows several real-world examples where the hand pose differs depending on object geometry. Overall, our experimental results across diverse daily objects demonstrate that \method is a promising approach for generalizable object pushing and pulling. It outperforms alternative methods such as querying the nearest hand pose in our data or using a fixed spatula-like hand pose, highlighting the need for a diffusion model to generate diverse hand poses. 

To summarize, the contributions of this paper include:
% \vspace{-6pt}
\begin{itemize}
    \item A scalable pipeline for generating and filtering dexterous hand poses for nonprehensile pushing and pulling. 
    \item A diffusion model for geometry-conditioned hand pose prediction for nonprehensile pushing and pulling. 
    \item A motion planning framework to execute these poses in the real world, with results across \numtrials trials showing that \method outperforms alternative methods. 
    \item A dataset of 1.3 million hand poses for nonprehensile pushing and pulling across 2.3k objects with corresponding canonical point cloud observations.
\end{itemize}

%\yunshuang{consider how to divide and describe a. dataset (least important) b. Method (DexGen+filtering) c. Long-horizon pushing} \daniel{(April 15) edited}

\begin{comment}
% Daniel: prior text from Yunshuang. Not sure if we will mention this but I tried to integrate it somewhat.
Uncertainty estimation during the non-prehensile manipulation — the non-prehensile manipulation is a dynamics process that there exists certain time range that is feasible for a grasp, eg. if keep pulling the book off the edge of the table, the book might fall off the table; if keep tilting the book then it might fall off the shelf. Uncertainty estimation during this dynamics process might be a key issue to increase the success rate.
\end{comment}

%===============================================================================

\section{Related Work}
\label{sec:related}

% Daniel: see DyWA paper for more to review.
% Daniel: mordatch2012contact is not nonprehensile, I would argue.
%\subsection{Nonprehensile Robot Manipulation}
\textbf{Nonprehensile Robot Manipulation.}
Classical nonprehensile manipulation includes pushing, sliding, rolling, and tilting, and has a long history in robotics~\cite{mason1986nonprehensile,lynch1996nonprehensile,mason1999nonprehensile,lynchmason1999nonprehensile}. 
Planning methods for nonprehensile manipulation often assume access to object models or priors~\cite{ChavanDafle2014dexterity,yang2024dynamiconpalmmanipulation}. Another recent planning-based method explores nonprehensile interaction with high-DOF hands in simulation by analyzing contact reasoning and wrench closure~\cite{Chen2023dexnonprehensile}. In contrast, our work targets real-world pushing and pulling using a high-DOF hand applied to diverse and geometrically complex objects. Applying traditional planning methods in this setting is highly challenging due to the complexity of modeling the diverse contact interactions between the hand and objects.
Recent learning-based methods have extended nonprehensile manipulation beyond classical planning, including extrinsic dexterity systems~\cite{zhou2022ungraspable,wu2024oneshottransferlonghorizonextrinsic} and those based on predicting object dynamics such as HACMan~\cite{zhou2023hacman,jiang2024hacmanpp}, CORN~\cite{cho2024corn}, and DyWA~\cite{lyu2025dywa}. 
Other works approach pushing as a precursor to grasping, often in planar settings with parallel-jaw grippers for multi-object manipulation~\cite{yonemaru2025learninggroupgraspmultiple}, or use bimanual systems for nonprehensile tasks using multi-link tools~\cite{liu2025factr}. 
None of these works study learning for single-hand pushing and pulling with dexterous hands. 
Furthermore, many prior benchmarks focus on pushing single flat objects on a surface, such as a T-shape object~\cite{chi2023diffusionpolicy}, or use spatulas to move small cubes and granular media~\cite{wang2023dynamicresolution,zhang2024adaptigraph}. 
Our work directly targets larger and more complex objects, including those that might topple or require coordinated multi-surface contact. 
%and we show how to use a dexterous robotic hand for nonprehensile manipulation. % Daniel: maybe not needed.

%Object pushing in particular has become a widely-used benchmark task, such as in the push-T task as tested in diffusion-based manipulation~\cite{chi2023diffusionpolicy}.  

%\subsection{Dexterous Grasping Synthesis and Datasets \daniel{TODO}}
\textbf{Dexterous Grasping Synthesis and Datasets.}
A substantial body of research focuses on generating and evaluating grasp poses for multi-fingered hands. Pioneering efforts such as~\cite{liu2020deepdifferentiablegraspplanner} create a dataset of 6.9K grasps using the GraspIt!~\cite{miller2004graspit} software tool, while~\cite{jiang2021hand} synthesize human hand poses by using a conditional Variational Autoencoder~\cite{kingma2022autoencodingvariationalbayes}. 
More recent efforts significantly scale grasp generation with tools such as differentiable contact simulation~\cite{turpin2022grasp,turpin2023fastgraspd} or optimization over an energy function based on Differentiable Force Closure (DFC)~\cite{Liu2022DFC}. 
Our work falls in the latter category, which has facilitated the generation of diverse grasping datasets such as DexGraspNet~\cite{wang2023dexgraspnet} with 1.32M grasps followed by DexGraspNet 2.0~\cite{zhang2024dexgraspnet} with 427M grasps, with recent work ~\cite{dexvlg25} further incorporating language into the dataset.
These pipelines generate hand poses by optimization over an energy function, filter them using physics simulators, train generative diffusion models for grasp synthesis, and typically include some fine-tuning or evaluation modules~\cite{weng2024dexdiffuser,lum2024gripmultifingergraspevaluation}. 
Most recently, Dex1B~\cite{ye2025dex1b} scales to one billion demonstrations by integrating optimization-based control with generative modeling.
While our pipeline also uses energy-based pose optimization and filtering, we focus on synthesizing hand poses for nonprehensile manipulation, specifically pushing and pulling, demonstrating the effectiveness of a tailored pipeline beyond grasping.

% Daniel: to cite:
%Others: Dexterous Functional Grasping~\cite{agarwal2023dexterous}, DVGG~\cite{wei2022dvggdeepvariationalgrasp}, datasets for multi-object grasping~\cite{li2024multigrasp,he2025sequential}. 

%\subsection{Learning-Based Dexterous Manipulation \daniel{TODO}}
\textbf{Learning-Based Dexterous Manipulation.}
Learning-based approaches for robotic grasping and manipulation have rapidly expanded in recent years~\cite{kroemer2021reviewrobotlearning}. While some recent work emphasizes fine-grained bimanual manipulation using parallel-jaw grippers~\cite{Zhao-RSS-23,fu2024mobile}, our focus is on learning single-arm manipulation with high-DOF dexterous hands such as the LEAP~\cite{shaw2023leaphand}, Allegro, and Shadow Hands. These hands have been applied to a variety of tasks, such as in-hand object rotation~\cite{qi2022hand,wang2024penspin,openai2019solvingrubikscuberobot}, object singulation~\cite{Hao2024SopeDex,xu2025dexsingrasp}, multi-object manipulation~\cite{he2025sequential,li2024multigrasp,yonemaru2025learninggroupgraspmultiple}, and bimanual systems~\cite{chen2025vegetables,lin2024twisting}.
While showing the versatility of dexterous hardware, these works focus on largely prehensile interactions. 
Prior learning-based systems with high-DOF hands for nonprehensile behaviors demonstrate tasks such as rolling objects or picking up plates as examples of learning from 3D data~\cite{Ze2024DP3} or human videos~\cite{lum2025crossinghumanrobotembodimentgap}. Recently,~\cite{chen2024task} synthesize task-oriented dexterous hand poses for certain nonprehensile tasks such as pulling drawers. However, none of these methods directly study pushing or pulling as their primary manipulation mode. %nor do they propose a pipeline for learning diverse pre-contact hand poses for executing nonprehensile manipulation. 
% Daniel: need to be careful about contrast with Chen et al.
    
%===============================================================================

\section{Problem Statement and Assumptions}
\label{sec:problem}

% Daniel: let's just present the basic problem here. 
%\yunshuang{observation assumption?}
We study nonprehensile object pushing and pulling on a flat surface using a single-arm robot with a high-DOF multi-finger dexterous hand (e.g., the Allegro Hand). By ``nonprehensile,'' we emphasize the distinction from ``prehensile pushing~\cite{perugini2025pushing}.'' We assume that there exists one object $O$ on the surface with configuration $S_{\rm obj} \in SE(3)$, and that the surface's friction properties facilitate object pushing. 
We use $P$ to indicate the object's point cloud sampled from its surface. 
%and the object's 3D position $t_{\rm obj}$ is its centroid \daniel{how do we do object positions?}. 
%and $B$ represent the corresponding \emph{basis point set}~\cite{prokudin2019BPS}.  % Daniel: maybe save BPS to later.
Let $\mathcal{H}$ be the space of possible pushing and pulling hand poses, where $H \in \mathcal{H}$ is defined as $H = (\theta, T)$. Here, $\theta \in \mathbb{R}^d$ is the joint configuration of the $d$-DOF robot hand, and $T \in SE(3)$ is the end-effector pose of the robot's wrist consisting of translation and orientation. 
A \emph{trial} is an instance of pushing or pulling, defined by a given direction $u_{\rm dir} \in \mathbb{R}^3$ (with z-component of 0) resulting in the target object position as $u_{\rm targ} \in \mathbb{R}^3$. 
The objective is to generate a hand pose $H$ such that, if a motion planner moves the hand to $H$ and then translates it along $u_{\rm dir}$, the object moves closer to the target $u_{\rm targ}$.  
The object's distance to $u_{\rm targ}$ must be below a threshold for a trial to be considered a success. %, while the hand maintains contact throughout.  % Daniel: maybe just mention the distance threshold, and we have more specific criteria later.
%This framework can be extended to multi-step pushing where multiple targets must be reached.

% Daniel: commenting out multi-object case, since even if we get that, we are going to use it just for a small qualitative example.
%$N_o$ objects, denoted as \(\mathbf{O} = \{ O_j \}_{j=1}^{N_o} \). 

%===============================================================================

\section{Method}
\label{sec:method}

\method consists of the following steps. First, we generate a large dataset of hand poses for pushing and pulling (Sec.~\ref{ssec:hand_poses}).
Second, we use this data to train a diffusion model to synthesize hand poses conditioned on object geometry (Sec.~\ref{ssec:diffusion_model}). 
Third, during deployment, we generate hand poses and perform motion planning (Sec.~\ref{ssec:motion_planning}). 

\subsection{Dataset Generation for Dexterous Pushing and Pulling}
\label{ssec:hand_poses}
\begin{figure}[t]
\center
\includegraphics[width=0.5\textwidth]{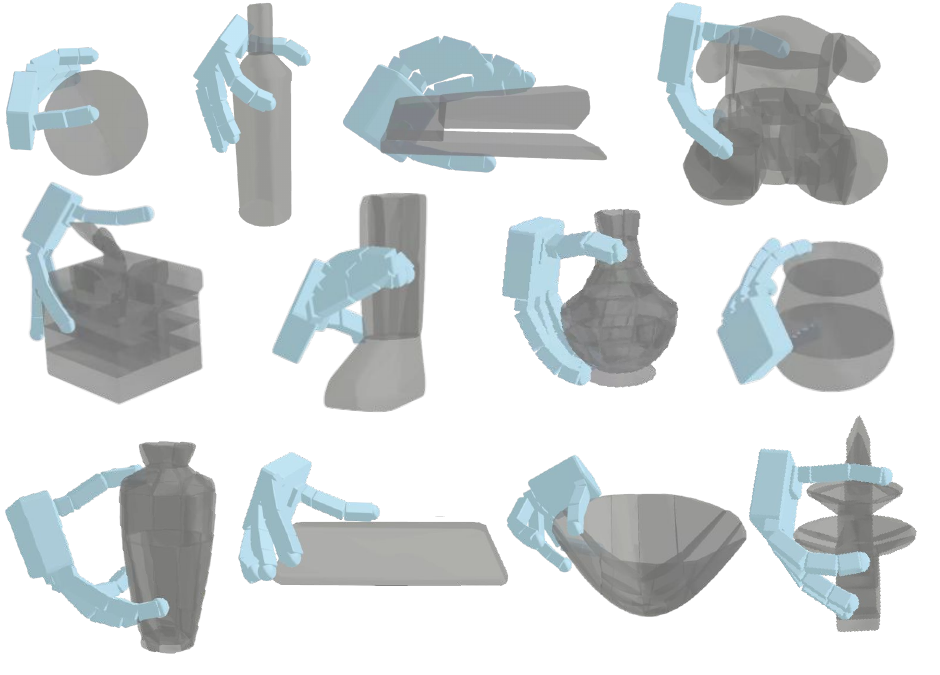} 
\caption{
Examples of pushing and pulling hand poses from optimizing our energy function (Eq.~\ref{eq:energy}). These have all been validated in IsaacGym simulation. In all examples, the intended object pushing direction is to the right. These data points are used to train our diffusion model (see Sec.~\ref{ssec:diffusion_model}). 
}
\label{fig:data_details}
\vspace*{-0.5cm}
\end{figure}  
We first generate diverse hand poses for pushing and pulling various objects in simulation. 
To do this, we take inspiration from prior work on generating diverse hand poses for \emph{grasping}~\cite{lum2024gripmultifingergraspevaluation,wang2023dexgraspnet,zhang2024dexgraspnet,xu2023unidexgrasp,li2024multigrasp,he2025sequential} by casting the hand synthesis problem as minimizing an energy function via optimization~\cite{Liu2022DFC}. Unlike those works, our focus is on pushing and pulling instead of grasping. 
To enable hand pose optimization, we define a set of candidate contact points sampled across the hand surface. Different regions of the hand have different candidate points to encourage broad contact across the palm and fingers. For the palm and finger (excluding fingertips), we sample points uniformly over the rigid body surface. For the fingertips, we sample from a denser set of points uniformly on the unit hemisphere for each tip. 
See the website for candidate contact points details. 
% (Figure~\ref{fig:contact_candidates} and Table~\ref{tab:contact_candidates}). % Daniel (April 28): did some shortening.

%\yunshuang{do we need to specify here}details the specific sampling regions and quantities.
%for a visualization of the parts of the hand considered for contact and the number of contact candidates for each embodiment region. 

With the sampled contact point candidates, we run an optimization algorithm following the sampling strategy from~\cite{lum2024gripmultifingergraspevaluation,wang2023dexgraspnet} that iteratively minimizes an energy function $E$ to generate hand poses. We adapt the energy function from~\cite{lum2024gripmultifingergraspevaluation} to suit our manipulation tasks, resulting in: 
\begin{equation}
    \label{eq:energy}
    E = E_{\rm fc} + w_{\rm dis} E_{\rm dis} + w_{\rm j} E_{\rm j} + w_{\rm pen} E_{\rm pen} + w_{\rm dir} E_{\rm dir} + w_{\rm arm} E_{\rm arm}
\end{equation}
where $E_{\rm fc}$ is a force closure estimator~\cite{Liu2022DFC}, $E_{\rm dis}$ penalizes hand-to-object distance (thus encouraging proximity),
$E_{\rm j}$ penalizes joint violations, and $E_{\rm pen}$ penalizes penetration between hand-object, hand-table and hand self-collision contacts. See~\cite{lum2024gripmultifingergraspevaluation,wang2023dexgraspnet} for further details. The $w$ terms are all scalar coefficients; we adopt the values from prior work and tune the weights (see the Appendix on our website) for the following two new terms. % from prior work, and we tune the other two that are new. 
To adapt the energy from Eq.~\ref{eq:energy} to pushing or pulling in a particular direction $u_{\rm dir}\in \mathbb{R}^3$, we introduce $E_{\rm dir}$ and $E_{\rm arm}$, which use the normal vector of the palm $v_{\rm palm} \in \mathbb{R}^3$. The $E_{\rm dir}$ term encourages $v_{\rm palm}$ to align with $u_{\rm dir}$,  and $E_{\rm arm}$ encourages hand poses that are kinematically feasible when attached to the robot arm. 
%Compared to the energy function designed in prior work~\cite{wang2023dexgraspnet,lum2024gripmultifingergraspevaluation}, we require key modifications on two aspects to tweak the hand poses proper for pushing (1) intentionally leverage palm to provide more stable contact during pushing or pulling with term $E_{\rm dir}$. (2) Pay more attention to the feasibility of kinematics solution of the hand poses when we attach it to a robot arm $E_{\rm arm}$. 
%We define a pushing direction as $p \in \mathbb{R}^3$ with z axis 0, the normal vector of palm as $v_{\rm palm} \in \mathbb{R}^3$. 
Formally, we define $E_{\rm dir}$ and $E_{\rm arm}$ as:
\begin{equation}
E_{\rm dir} \;=-\;
\frac{u_{\rm dir}^{\mathsf T}v_{\rm palm}}{\|u_{\rm dir}\|_2 \| v_{\rm palm}\|_2} 
%\;
\mbox{;}
\;\;
E_{\rm arm} = \max \bigl( 0, (v_{\rm palm})_z \bigr)
% E_{\rm arm} =
% \begin{cases}
% \bigl(v_{\rm palm})_{z} & \text{if }\bigl(v_{\rm palm})_{z} \geq 0,\\[6pt]
% 0 & \text{otherwise }
% \end{cases},
\end{equation}
where $(v_{\rm palm})_{z}$ is the $z$-component of the palm's normal vector (in the world frame). Intuitively, aligning $u_{\rm dir}$ and $v_{\rm palm}$ promotes more stable object-palm directional contact. Furthermore, if the palm faces upwards, then the rest of the arm must be below it. Thus, it is likely to lead to an infeasible robot configuration due to robot-table intersections, so $E_{\rm arm}$ is nonzero (i.e., worse). To inject randomness (and thus diversity) in the sampling process, we randomly resample a subset of the contact point indices from the set of valid candidates 
% (Figure~\ref{fig:contact_candidates}) 
when generating a new hand pose. We use RMSProp~\cite{tieleman2012rmsprop} to update translation, rotation and joint angles with step size decay, then minimize the energy function with Simulated Annealing~\cite{kirkpatrick1983optimization} to adjust parameters. 

%\subsection{Hand Pose Validation in Simulation} 
\textbf{Hand Pose Validation in Simulation.} 
After optimizing contact points to generate candidate hand poses, we must \emph{validate} whether they can lead to successful pushing or pulling. To do this, we use IsaacGym~\cite{makoviychuk2021isaac}, a GPU-accelerated physics simulator that has been used in prior work for filtering grasp poses~\cite{lum2024gripmultifingergraspevaluation,wang2023dexgraspnet}. 
%We filter hand poses for nonprehensile manipulation, and 
We define a push or a pull as successful if, after executing a \SI{20}{\centi\meter} translation, the object's center is within \SI{3}{\centi\meter} of the target position \emph{and} the object's orientation changes by no more than 45 degrees relative to its original configuration. 
%push/pull ``success'' as follows:\yunshuang{please help me polish} after execution the pushing/pulling with \SI{20}{\centi\meter}, the objects is within \SI{3}{\centi\meter} threshold to the target and within 45 degree changes to its original configuration. 
The optimization process has a low success rate because it does not account for the full dynamics of pushing and pulling. Thus, we augment successful hand poses by adding slight noise to the pose parameters. From extensive parallel experiments, we generate a dataset containing 2.3k objects with 1.3 million successful hand poses.  See the Appendix on our website for more details. 
%\yunshuang{refer to figures in append for details on data generation and statistical analysis on dataset}

\subsection{Training a Diffusion Model to Predict Hand Poses}
\label{ssec:diffusion_model}

To generate hand poses, we adapt a conditional U-Net~\cite{ronneberger2015unet} from the diffusion policy architecture~\cite{chi2023diffusionpolicy}, and train it with the Denoising Diffusion Probabilistic Models (DDPM) objective~\cite{ho2020denoisingdiffusionprobabilisticmodels}. 
Diffusion models are well-suited for this task as they can learn complex, high-dimensional distributions. The forward process gradually adds Gaussian noise to the hand configuration $H$, while the reverse process reconstructs the original pose $H$ by iteratively denoising conditioned on the object's geometry. 
% \begin{equation}
% p_{\phi}(H_{t-1}|H_t, B) = \mathcal{N}(
%       H_{t-1};
%       \mu_{\phi}(H_{t}, t, B),
%       \Sigma_{\phi}(H_{t}, t, B)
%     \bigr).
% \end{equation}
% The input \daniel{the hand pose is output, not input?} is the hand poses $H = (\theta, T) \in \mathbb{R}^{25}$, including 3D hand position, 6D orientation relative to the object frame, and the 16 joint angles of the Allegro Hand. We randomly sample a time step $t$, then apply the DDPM scheduler to add noise to the input hand poses, conditioned on the object observation.
The model is trained to minimize denoising error. To represent the observation, we use a 4096-dimensional Basis Point Set (BPS)~\cite{prokudin2019BPS} representation $B \in \mathbb{R}^{4096}$ based on the object's point cloud $P$. 
For 3D-printed objects, we use their known meshes to directly compute their BPS representation. For the other objects, we follow the pipeline proposed in~\cite{lou2024robo} to obtain real-world object point clouds (and thus, the BPS).
Specifically, we reconstruct object meshes by using Nerfstudio~\cite{nerfstudio} and we
% to compute COLMAP reconstructions~\cite{structureFromMotion2016}. We 
use Stable Normal~\cite{ye2024stablenormalreducingdiffusionvariance} to generate normal maps. Then, we employ 2D Gaussian Splatting~\cite{Huang_2024} to obtain the point clouds.
This representation, which is also used in~\cite{lum2024gripmultifingergraspevaluation,weng2024dexdiffuser}, encodes each object as a fixed-length vector of shortest distances between canonical basis points and the points in $P$. BPS captures geometric properties in a compact manner and simplifies the design of the diffusion model. %with points that are a fixed length while also capturing appropriate shape properties compared to point clouds.  It represents the shortest distance between each point in a fixed set of basis points and the points in $P$. 
Given this trained diffusion model, at test time it can be used to generate diverse hand poses which we can select for motion planning. 
See Fig.~\ref{fig:method} and Appendix
% ~\ref{ssec:app_training_details} 
on our website for more information.

\subsection{Arm-Hand Motion Planning and Evaluation} 
\label{ssec:motion_planning}

\begin{figure}
  \centering
  \includegraphics[width=\linewidth]{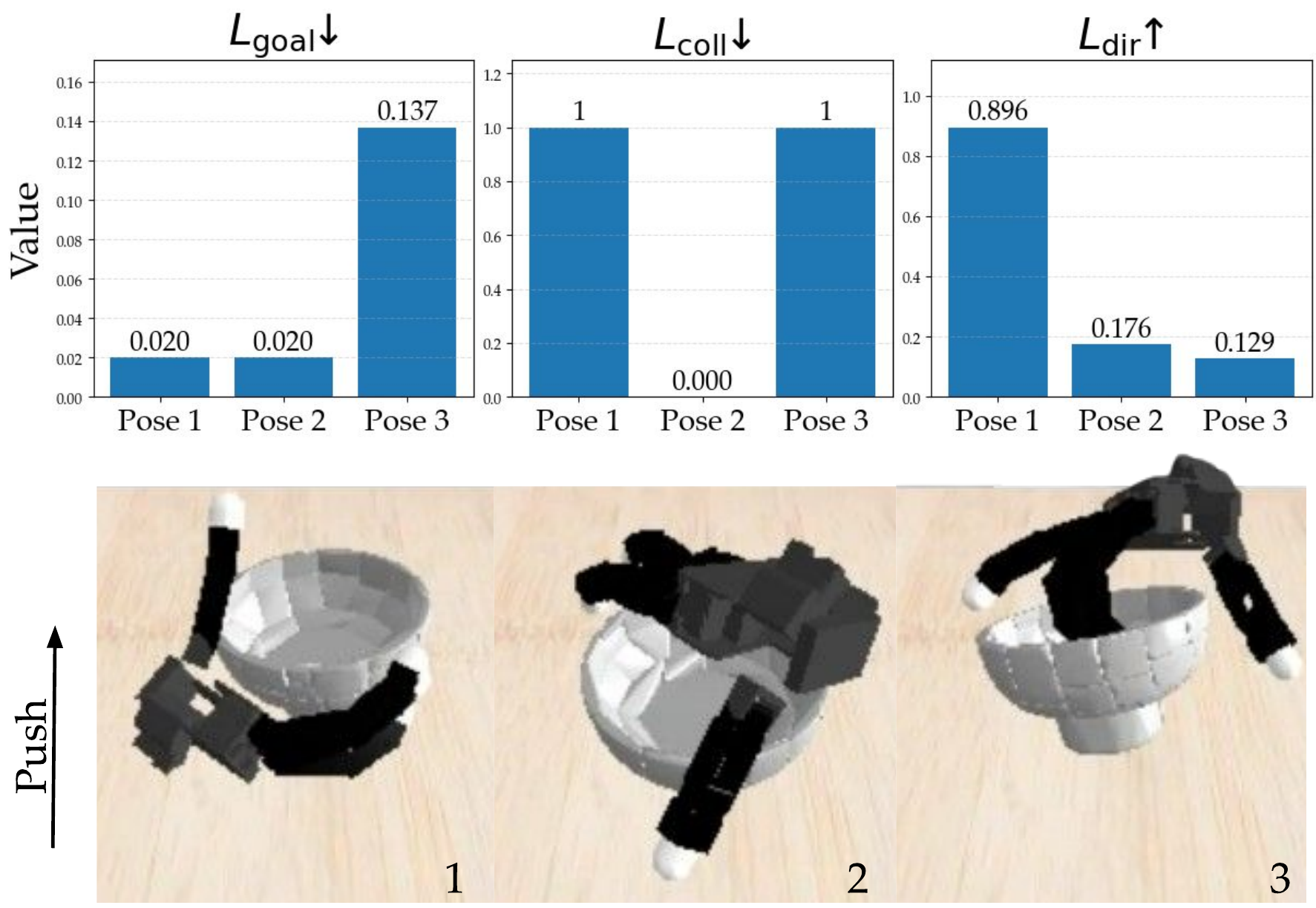}
  \caption{
  Visualization of $L_{\rm goal}, L_{\rm coll}$, and $L_{\rm dir}$ values in $V(H)$ from Eq.~\ref{eq:eval_metric} on three simulated hand poses. See Sec.~\ref{ssec:motion_planning} and Sec.~\ref{ssec:real_results} for more details.
  % Poses 1 \& 2 have similar positional success, but Pose 1 collides with the table, making Pose 2 preferable under our ranking.
  }
  \label{fig:ablation}
\vspace*{-10pt}
\end{figure}

During deployment, the diffusion model generates candidate hand poses. We then integrate the robotic arm into full arm-hand motion planning to select hand poses which are kinematically feasible and avoid environment collisions, such as arm-table intersections (which are not considered in Sec.~\ref{ssec:hand_poses}). 
%This will now integrate the Franka arm into the process so that we can simulate things like hand-arm feasibility and if the arm would intersect with the table. By testing rollouts in simulation, we can perform additional checking and verification and rank the rollouts (and thus, hand poses). 
See Fig.~\ref{fig:method} (right half) for an overview. 
Each hand pose $H = (\theta, T)$ is initially expressed in the object frame. We use the object's initial configuration $S_{\rm obj}$ and intended direction $u_{\rm dir}$ to transform $H$ to the world frame, and supply that to the cuRobo planner~\cite{sundaralingam2023curobo} to generate a complete motion plan for the robotic arm. 
%For each hand pose $H = (\theta, T)$, we transform $H$ in the world frame given the $u_{\rm targ}$ and $u_{\rm dir}$ of the corresponding object and input to cuRobo~\cite{sundaralingam2023curobo} motion planning framework. 
%This generates a complete motion plan for the Franka arm with the hand to execute the pushing or pulling action. 
In this process, we discard infeasible trajectories (and thus, the associated hand poses) to only keep the feasible arm-hand trajectories. To select which of the feasible trajectories to execute, we associate each with a custom analytical score $V$, defined as:
%For planning arm and hand motions, we input the target $H(\theta, T)$, object configuration $S_{\rm obj} \in SE(3)$, and the pushing instance $(\phi, d)$ ($\phi$: pushing direction, $d$ pushing distance) into cuRobo~\cite{sundaralingam2023curobo} to plan the whole arm and hand action sequence over the pushing instance. Therefore, we obtain intact hand and Franka arm motions corresponding to each generated hand pose with a score. Then during deployment, we execute the one with the highest score that potentially ensures best pushing performance with least possibility to collide in the real world.
%Given all the rollouts, we define an analytical evaluation metric $V$ to rank all the inference hand poses with
\begin{equation}
\label{eq:eval_metric}
V(H = (\theta, T)) = \alpha L_{\rm goal} + \beta L_{\rm coll} + \gamma L_{\rm dir},
\end{equation}
where $L_{\rm goal}$ measures the Euclidean distance between the object's final position and the target position, $L_{\rm coll}$ indicates whether a collision occurred during execution (1 if a collision occurs, 0 otherwise), and $L_{\rm dir}$ encourages the palm's orientation to align with the pushing direction. For $L_{\rm dir}$, we set it equal to the $E_{\rm dir}$ term from the energy function (Eq.~\ref{eq:energy}).  
The $\alpha, \beta$, and $\gamma$ are hyperparameters tuned empirically.

To provide additional context, Fig.~\ref{fig:ablation} presents three candidate hand poses along with their corresponding scores on $L_{\rm goal}$, $L_{\rm coll}$ (where lower values are preferable), and $L_{\rm dir}$ (where higher values are preferable). Hand pose 1 achieves the highest score on $L_{\rm dir}$, indicating effective alignment with the pushing direction, yet it results in an avoidable hand–table collision. Hand pose 3 exhibits a low score on $L_{\rm goal}$, suggesting limited effectiveness in pushing the object toward the target. Balancing all three objectives, the motion planner would select hand pose 2 for execution in the real world. We tune these weight terms according to their influence on pushing feasibility and overall task success.

\textbf{Multi-step Planning.} While we mainly study \method for single open-loop pushes (or pulls) to targets, our framework naturally extends to multi-step planning. 
In scenarios with obstacles, we first compute a collision-free global path using RRT*~\cite{karaman2011sampling}. Then, we sequentially plan hand poses to reach each intermediate waypoint. 
Given an object, the same hand pose may be feasible only in certain pushing or pulling directions due to robot and hand kinematics. %We only incorporate feasible motions into the multi-step plan. 
The waypoints from RRT* may require planning pushes across challenging directions, highlighting the importance of generating diverse hand poses for varying object positions and directions. 
%InHowever, we can extend this to a multi-step system by planning for multiple objectives and then re-planning once we achieve different waypoints. In scenarios with obstacles, we first do path planning with RRT*~\cite{karaman2011sampling}. Then for each waypoint, we plan for a proper hand pose to execute consecutive pushing or pulling along the path to avoid collision with the obstacle. This motivates our objectives on pushing directions since we have to generate hand poses under different position with different directions, which primarily is difficult for the kinematics of the hand and arm setting. 

%===============================================================================

\begin{comment}
\begin{figure*}[t]
\center
\includegraphics[width=1.0\textwidth]{figures/setup_objects_v6.pdf} 
\caption{
Left: our physical setup using a Franka and an Allegro Hand, along with a flat workspace. 
%\daniel{the camera is going to confuse readers as we don't use it} 
Right: the objects we use for real-world nonprehensile manipulation experiments. We test with 3D printed objects as well as common (``Daily'') everyday objects. %\daniel{Label / index objects?} \yunshuang{How can we specify the object somewhere}
See Sec.~\ref{ssec:real_experiments} for more details. 
} 
\label{fig:object_vis}
\vspace*{-10pt}
\end{figure*}  
\end{comment}

\begin{figure}
    \centering
    \includegraphics[width=1\linewidth]{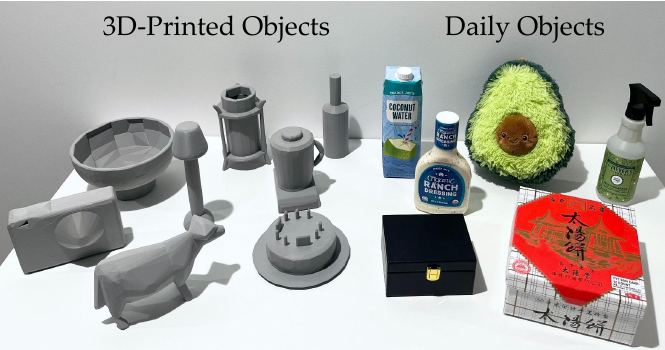}
    \caption{  %Left: our physical setup using a Franka and an Allegro Hand, along with a flat workspace. 
  The objects we use in our real-world experiments, including 3D printed and daily objects. See Sec.~\ref{ssec:real_experiments} for more details.}
    \label{fig:object_vis}
\vspace*{-10pt}
\end{figure}

\section{Experiments}
\label{sec:experiments}

%Our method, \method, generates diverse dexterous hand poses for objects with varying geometries, enabling nonprehensile manipulation.  
Through simulation and real-world experiments, we aim to investigate the following questions: 
(1) Can we learn feasible and effective hand poses from our large-scale dataset? 
(2) Is \method more effective and robust  for dexterous pushing and pulling compared to baselines and ablations?
(3) Can \method serve as a reliable module for downstream manipulation tasks such as multi-step pushing around obstacles?
(4) Does \method perform consistently across different hand morphologies?

\subsection{Simulation Experiments and Results}
\label{ssec:simulation_experiments}

We investigate the first question in simulation.
We evaluate the quality of the hand pose generation pipeline using IsaacGym~\cite{makoviychuk2021isaac}. To quantify the effectiveness of our 
\begin{wraptable}{r}{0.29\textwidth}
%\begin{table*}[t]
  \vspace{-6pt}
  \setlength\tabcolsep{4.6pt}
  \centering
    \footnotesize
    \begin{tabular}{cc}
    \toprule
    Data Size
    & \multicolumn{1}{c}{\# of Objects}  \\ 
    \midrule
      2\% &  41.67 $\pm$ 10.21  \\ 
     20\% & 102.67 $\pm$ \;5.85 \\
     50\% & 110.33 $\pm$ 29.67  \\
    100\% & 169.33 $\pm$ 15.18  \\ 
    \bottomrule
    \end{tabular}
  \caption{Number of objects with $\ge 1$ feasible pushing hand pose out of 300.}
  % \vspace*{-15pt}
  \label{tab:sim-results}
%\end{table*}  
\end{wraptable}
trained model and dataset, we report the number of successfully pushed objects as a function of training data size. 
We train our diffusion model on varying subsets of the full dataset (of 1.3M hand poses) and evaluate on 300 unseen objects from the test set. The objects we use for training and evaluation come from ~\cite{lum2024gripmultifingergraspevaluation}, consisting of diverse daily objects. For each test object, we sample 200 candidate hand poses. An object is considered ``successful'' if at least one feasible hand pose results in success. 
Table~\ref{tab:sim-results} reports results over 3 different seeds, %which shows that performance improves with dataset size. 
%Quantitatively, we observe our trained model is able to generate reasonable pushing poses with various object geometries more reliably as the dataset size grows, which supports the value of having such scale of dataset for pushing objects. However, it doesn't grow exponentially as expected. We believe train more powerful model with proper tuning might result in a better statistical data. 
%We would leave that as a future work and encourage the community to explore the topic. 
which shows that our model generates feasible pushing poses more reliably with larger training sets, which validates large-scale supervision. The growth is not strictly linear, suggesting room for improvement via better model tuning or data strategies. 
Qualitatively, our generated hand poses are diverse across object geometries and exhibit pushing intent (see the Appendix on our website for more discussion). A common failure mode in the preliminary experiments is that some generated poses still collide with the object, which motivates the later inclusion of the collision term in Eq.~\ref{eq:eval_metric} for motion planning to further avoid collisions before executing pushing and pulling motions in the real world. 
%when executing pushing motion in real world experiments.

\subsection{Real-World Experiments Setup}
\label{ssec:real_experiments}

\begin{figure*}[t]
\center
\includegraphics[width=1.0\textwidth]{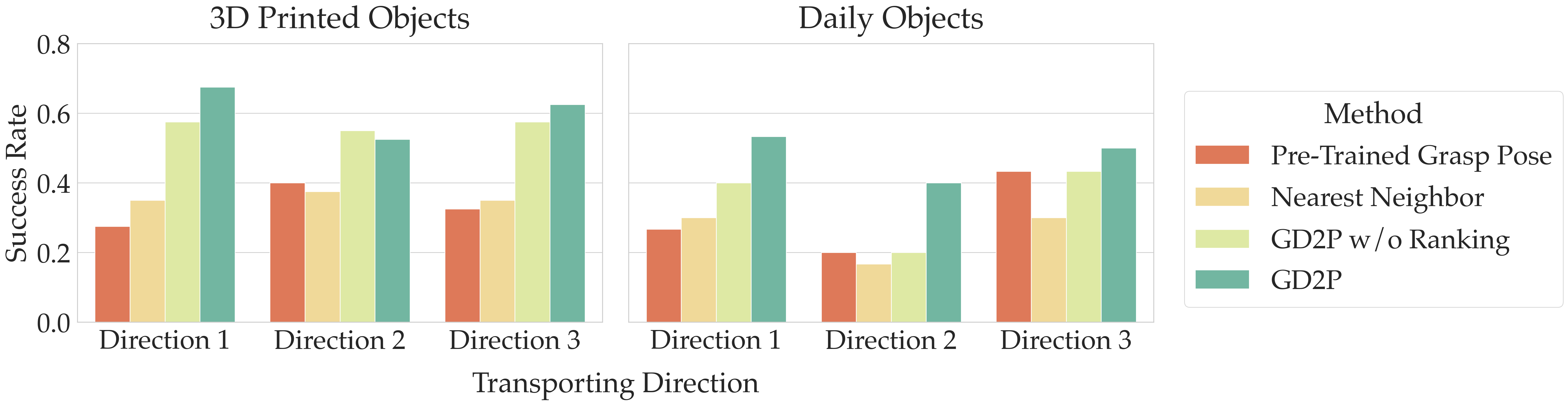} 
\caption{ Pushing and pulling success rates from \method and baselines, across different 3D printed (left) and daily objects (right), and with three directions evaluated. Each bar aggregates success rates from 40 trials (left bar plot) and 30 trials (right bar plot). 
See Sec.~\ref{ssec:real_experiments} and~\ref{ssec:real_results} for more details.
}
\label{fig:bar_plot_baselines}
\vspace*{-5pt}
\end{figure*}  

 %~\cite{nerfstudio} 
%Computes COLMAP reconstructions~\cite{schoenberger2016sfm} and Stable Normal~\cite{ye2024stablenormal} to generate normal maps. 
%Then 2D Gaussian Splatting~\cite{Huang2DGS2024} to obtain the point clouds. 
%To determine the camera's pose relative to the scene, we use AprilTag markers~\cite{olson2011tags}. 

We evaluate \method on a real robot to check if our pushing and pulling hand poses successfully transfer to reality. Our hardware setup consists of a Franka Panda arm equipped with a four-finger, 16-DOF Allegro Hand (see the Appendix on our website for detailed setup visualization). 
%It operates over a tabletop High Density Polyethylene Sheet cutting board with dimensions \SI{60}{\centi\meter}$\times$\SI{60}{\centi\meter}.
It operates over a tabletop cutting board with dimensions \SI{60}{\centi\meter}$\times$\SI{60}{\centi\meter}.
We use a mix of objects, including 3D-printed and common daily items (shown in Fig.~\ref{fig:object_vis}). All evaluation objects are unseen during training. We obtain the object BPS representation as stated in~\ref{ssec:diffusion_model}. While this pipeline introduces some noise, it is sufficient for \method to predict effective hand poses. In contrast, we empirically observed that optimization-based methods are more sensitive to mesh quality and often fail under these conditions. %This highlights how \method can apply using only moderately accurate reconstructions, without requiring highly precise meshes.
% We introduce this visual representation to show the robustness of our method to different visual representations and it indicates that our method could work on any objects (we assume we could always get the mesh from off-the-shelf method \daniel{add to assumptions section}).

%\subsubsection{Baselines and Ablations} 
\textbf{Baselines and Ablations.} 
We compare \method with:
\begin{itemize}
\item \textbf{Pre-Trained Grasp Pose}: We use a pre-trained grasp synthesis model from~\cite{lum2024gripmultifingergraspevaluation} using NeRF~\cite{mildenhall2020nerf}. For each object, we train a NeRF representation, then query their pre-trained model for a grasp. This evaluates how well a grasping-centric model generalizes to nonprehensile tasks. 
\item \textbf{Nearest Neighbor (NN)}: Given a test object, we find the training object with the most similar BPS representation (in terms of Euclidean distance) and retrieve its associated hand poses. We then do the same motion planning pipeline as in \method. This tests out-of-distribution generalization with a retrieval-only approach compared to our proposed generative model.
\item \textbf{\method w/o Ranking}: An ablation that excludes analytical ranking of hand poses (ignores Eq.~\ref{eq:eval_metric})  and executes a random feasible pose. This tests the usefulness of Eq.~\ref{eq:eval_metric} in selecting poses. % our post-generation evaluation helps select a more promising hand pose.  
\end{itemize}
% \vspace*{-3pt}

%\subsubsection{Experiment Protocol and Evaluation}
\textbf{Experiment Protocol and Evaluation.}
% from yunshuang: when I did the motion planning for the arm+hand, they would end up at the goal position. So if the hand and objects are still in contact then I count it as a success
%We evaluate the success rate across multiple directions and trials. 
For each object, we test three pushing and pulling directions uniformly distributed around a circle. We consider ``pulling" to involve applying force towards the body to move an object. This aligns with ``Direction 2'' in our real-world experiments. Along each direction, the robot executes the hand pose and planned motion five times, all with a fixed push length of \SI{20}{\centi\meter}. 
A human manually places the object in a relatively consistent pose between trials. A trial is successful if the object's center is within \SI{3}{\centi\meter} of the target position, the hand maintains contact throughout, and it does not lead to task failure modes such as toppling or loss of control.  %the push does not with the object and it does not cause major object damage (e.g. toppling).
For NN and \method w/o Ranking, we randomly sample hand poses among the feasible planned actions. For Pre-trained Grasp Pose, we execute the best actions from its output. For our method, we execute the one with the highest score from Eq.~\ref{eq:eval_metric}. 
%\daniel{anything else to add? What if it maintains contact but topples? Also maybe state clearly how many total trials are run to reiterate the number in abstract.}
\begin{figure}[h]
\center
\includegraphics[width=1.0\linewidth]{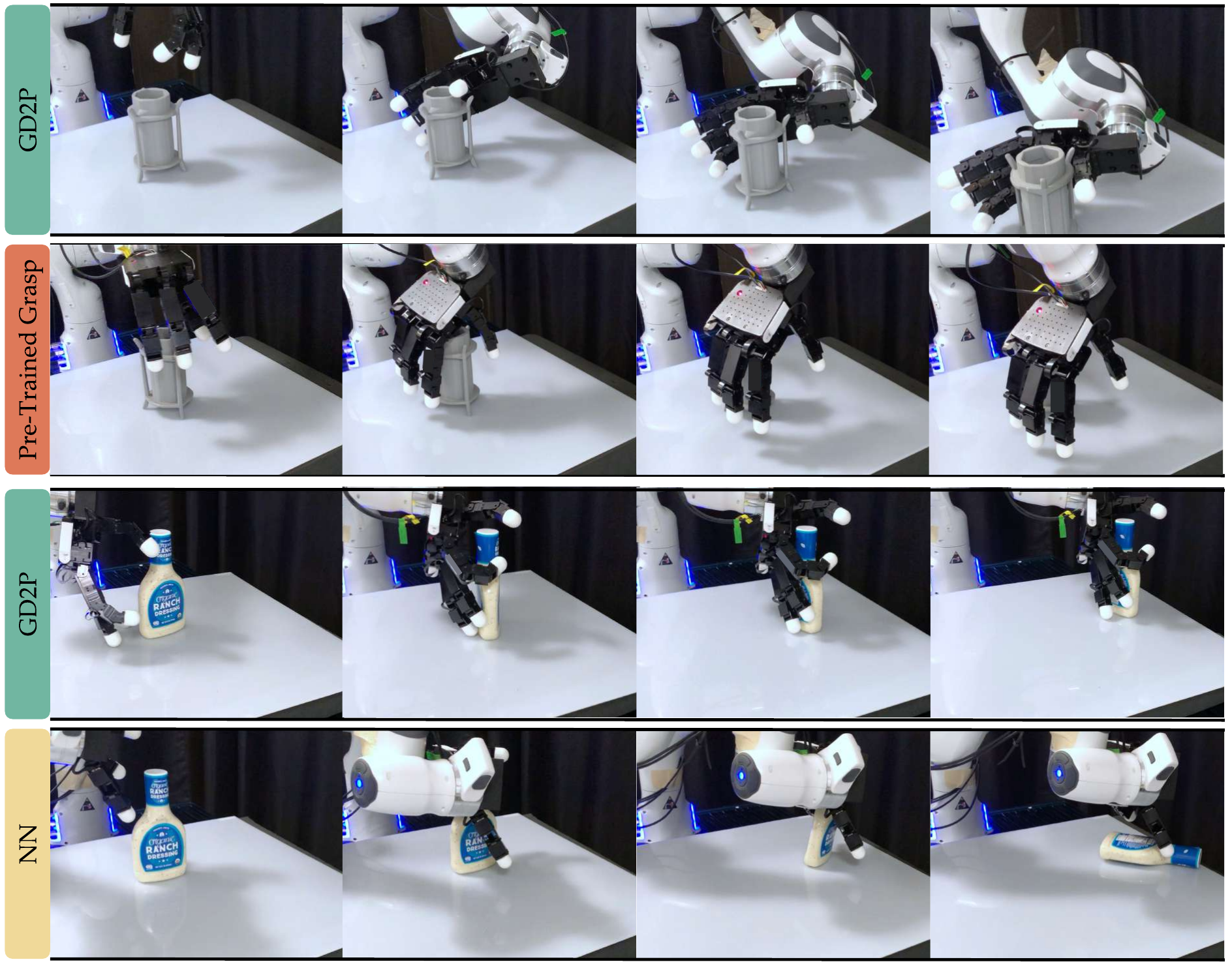} 
\caption{
Comparison between \method and baselines. The first two rows show \method (success) and Pre-Trained Grasp (failure) while pushing a 3D-printed vase forward (i.e., away from the robot). The last two rows show \method (success) and NN (failure) while pushing a ranch bottle to the right.}
\label{fig:results_real}
\vspace*{-10pt}
\end{figure} 

\begin{figure}[t]
  \centering
  \includegraphics[width=\linewidth]{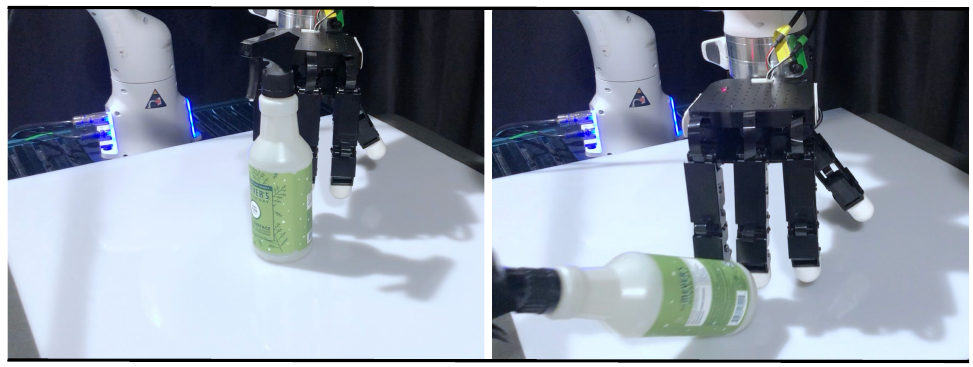}
  \caption{
  Example of a typical failure case using the Fixed Hand Pose strategy (see Sec.~\ref{ssec:real_results} for details), which topples the spray.
  }
  \label{fig:fix_pose}
\vspace*{-10pt}
\end{figure}

% \begin{figure*}[t]
% \center
% \includegraphics[width=0.5\textwidth]{figures/multistage_v01.pdf} 
% \caption{
% \yunshuang{Showing scenarios of multi-step pushes and different poses}
% }
% \label{fig:results_multistep}
% \vspace*{-5pt}
% \end{figure*}  

\begin{comment}
%\begin{wraptable}{r}{0.73\textwidth}
\begin{table*}[b]
  \setlength\tabcolsep{4.6pt}
  \centering
    \footnotesize
    \begin{tabular}{lcccccccccc}
    \toprule
    & \multicolumn{1}{c}{Easy} & \multicolumn{1}{c}{Medium} & \multicolumn{1}{c}{Hard} \\ 
    Method & \multicolumn{1}{c}{(3 Objects)} & \multicolumn{1}{c}{(3 Object)} & \multicolumn{1}{c}{(3 Objects)} \\ 
    \midrule
    kNN & ... & ... & ... \\ 
    Fixed Hand Poses & ... & ... & ... \\ 
    Get a Grip Pre-Trained~\cite{lum2024gripmultifingergraspevaluation}  & ... & ... & ... \\ 
    \method w/o estimation & ... & ... & ... \\ 
    \rowcolor{lightgreen} \method (ours) & ... & ... & ... \\ 
    \bottomrule
    \end{tabular}
  \caption{
    \daniel{Real-World Experiments. TODO Yunshuang fill this in, e.g., with values like ``4/5'' depending on results, etc. Also, we can split the columns into ``Easy - 3D printed,'' ``Easy - Real'' etc....}
  }
  \vspace*{-5pt}
  \label{tab:real-results}
\end{table*}  
%\end{wraptable}
\end{comment}

\subsection{Is \method Effective for Diverse Objects and Directions?}
\label{ssec:real_results}
We summarize quantitative results in Fig.~\ref{fig:bar_plot_baselines}, which shows that \method outperforms or matches alternative methods for both object categories. 
As shown in Fig.~\ref{fig:results_real}, the \textbf{Pre-Trained Grasp Pose} baseline suffers from two major issues. First, the hand pose is not conditioned on the pushing direction, which means during the push, the object is likely to slide off the hand due to limited support (Fig.~\ref{fig:results_real}, second row). Second, some objects are unsuitable for grasping with a single hand due to their geometry or awkward aspect ratios (e.g., a flat box with limited area for enclosure). 
Additionally, the similarity-based \textbf{Nearest Neighbor} baseline struggles due to limited granularity in object geometry matching, motivating the need for our geometry-conditioned generative model.
For \textbf{\method w/o ranking}, we observe that its hand poses are more likely to collide with the table or objects. 
% \begin{wrapfigure}{r}{0.5\textwidth}
% % \vspace{-5pt}
% \includegraphics[width=0.5\textwidth]{figures/multistage_v02.pdf} 
% \caption{
% Two examples of multi-step pushes using \method, which both avoid the central obstacle. 
% %\yunshuang{Showing scenarios of multi-step pushes and different poses}
% }
% \label{fig:results_multistep}
% % \vspace*{-5pt}
% \end{wrapfigure}
To further investigate this ablation, Fig.~\ref{fig:ablation} shows three different hand poses. 
The first one has a low collision score because it is easy to collide with the table, while the third collides with the objects and scores low on the palm direction. 
The second hand pose leads to a successful push in real-world experiments. This suggests the importance of our ranking system via Eq.~\ref{eq:eval_metric}. 
\textbf{\method} outperforms baselines in all directions tested in Fig.~\ref{fig:bar_plot_baselines}, demonstrating the robustness of its generated hand poses for pushing and pulling objects. Fig.~\ref{fig:results_real} (first row) demonstrates using the palm and thumb to provide strong support moving the object forward, and the third row shows using the thumb and index finger to form a circular shape support for the thinner upper parts of the object while providing force at the bottom, aiding stable movement. For more rollouts, see the Appendix and the website.

% \begin{figure}
%   \centering
% \includegraphics[width=\linewidth]{figures/multistage_v03.pdf}
%   \caption{
%    Example of multi-step pushes using \method, which avoids the central obstacle. 
%   }
%   \label{fig:results_multistep}
% \end{figure}

\begin{figure}[t]
\centering
\includegraphics[width=\linewidth]{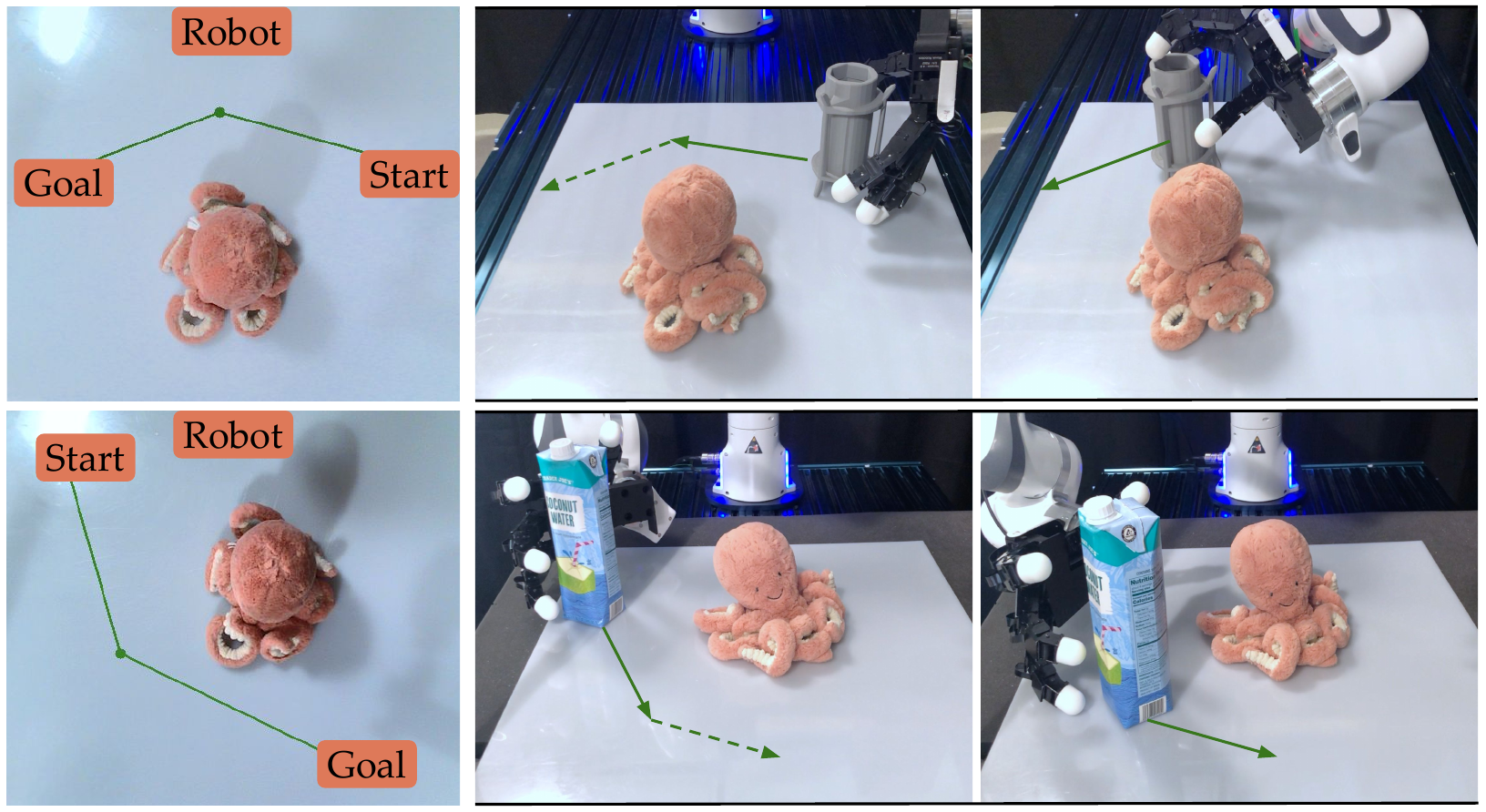}
\caption{
Path planning using RRT* for multi-step planning. The first column shows the visualization of path planning results. The second and third columns show two consecutive hand poses for pushing the object along the path.
}
\label{fig:results_multistep}
\vspace*{-10pt}
\end{figure}

\emph{Fixed Hand Pose}: Inspired by prior pushing work~\cite{wang2023dynamicresolution}, we manually define a ``spatula'' hand pose with the fingers spread flat (see Fig.~\ref{fig:fix_pose}) to assess whether simple flat-hand strategies suffice for diverse objects. %This strategy is inspired by prior pushing work~\cite{wang2023dynamicresolution}. 
We perform a case study on the 6 objects in Fig.~\ref{fig:object_vis} that are taller than \SI{20}{\centi\meter}. We push each object 10 times
% , with 5 pushes for each of 2 directions, (the third direction results in kinematic errors). 
and get a relatively low 18/60 success rate, suggesting insufficient object support.

\subsection{Can \method Support Multi-step Manipulation Tasks?}
\textbf{\method can serve as a reliable module for multi-step pushing.} 
Selecting a kinematically feasible hand pose for a given object state $S_{\rm obj}$ and direction $u_{\rm dir}$ is challenging in multi-step planning, as different waypoints may require different hand poses.
%The challenge of multi-step planning lies in selecting a kinematically feasible hand pose for a given $S_{\rm obj}$ and $u_{\rm dir}$, as it may result in different hand poses selection along the way. 
Our method resolves this by identifying valid poses across object configurations and coupling pose selection with kinematic feasibility (see Sec.~\ref{ssec:motion_planning}). By doing so, \method can be used to perform multiple pushes. % (see Sec.~\ref{ssec:motion_planning}).  
Fig.~\ref{fig:results_multistep} shows a multi-step pushing sequence using \method. The robot uses two different hand poses to push the 3D-printed vase, as the first hand pose may not be ideal for the second hand pose, which shows the benefit of re-planning. 
%Figure~\ref{fig:results_multistep} shows two multi-step pushing sequences using \method. In the first, the robot uses two different hand poses to push the 3D-printed vase, as the first hand pose may not be ideal for the second hand pose, which shows the benefit of re-planning. In the second example, the robot can use a similar hand pose for both pushes. 

\subsection{Does \method Apply to Different Hand Morphologies?}

\begin{figure}
  \centering
\includegraphics[width=\linewidth]{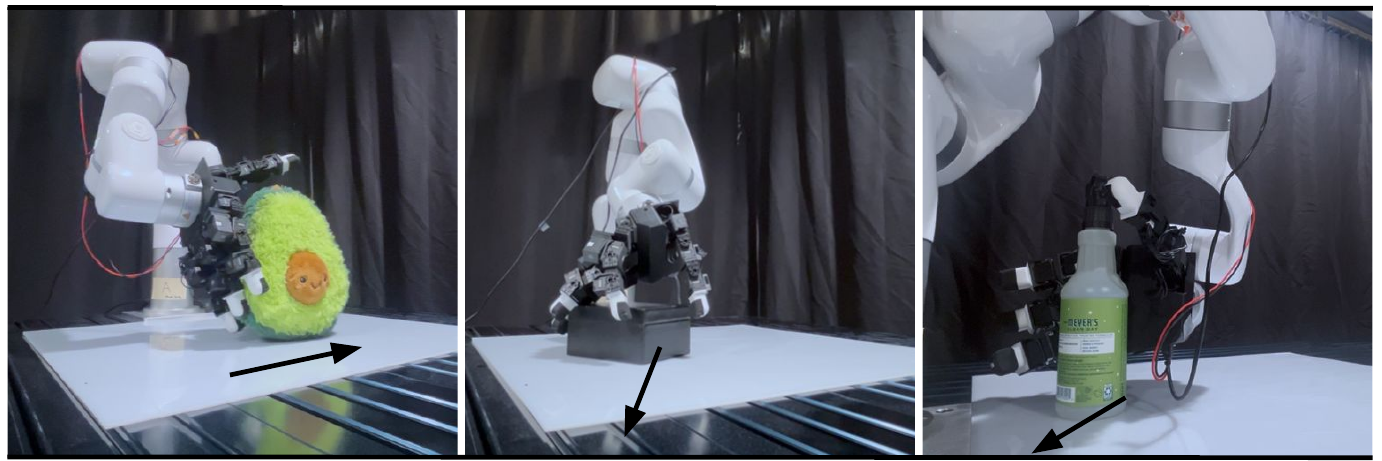}
  \caption{
  Pushing and pulling experiments with a LEAP Hand, following the same protocol and evaluations from Section~\ref{ssec:real_experiments}. We evaluates using the same 14 objects across three directions. The first two images illustrate pushing motions and the third illustrates a pulling motion of the spray bottle toward the robot base.
  }
  \label{fig:leap}
  \vspace*{-10pt}
\end{figure}

\textbf{\method achieves consistent performance across various robotic hand designs.} To evaluate the robustness of \method across different robotic hands, we applied the same pipeline on a four-finger, 16-DOF LEAP Hand~\cite{shaw2023leaphand} with an xArm7 arm and conducted experiments following the procedure described in Section~\ref{ssec:real_experiments} (see Fig.~\ref{fig:leap}). Specifically, we tested on the same set of 14 objects over three directions in 210 new trials (not part of the \numtrials trials in Sec.~\ref {ssec:real_results}) and achieved a comparable success rate of 68.1\%. These results highlight the potential of \method for generating effective dexterous pushing and pulling policies across different robotic hands.

\begin{comment}
\begin{figure*}[t]
\center
\includegraphics[width=0.5\textwidth]{figures/ranking_plot_v01.pdf} 
\caption{Visualization of breakdown $V(H)$ from Eq.~\ref{eq:eval_metric} (unweighted) on three hand poses shown in IsaacGym simulation. Poses 1 \& 2 have comparable performance in terms of reaching the goal position. However, 1 has collisions with the table, leading to the final choice of 2 under our ranking criterion.
}
\label{fig:ablation}
\vspace*{-5pt}
\end{figure*}  
\end{comment}

% Qualitatively, \method is able to lead to dexterous pushing where some objects might be at risk of toppling (e.g., ranch bottle), which distinguishes our work from most in the ``pushing'' literature which tend to use objects that are flatter such as cubes or granular media~\cite{wang2023dynamicresolution,zhang2024adaptigraph} which do not topple. Figure~\ref{fig:results_real} shows some qualitative results between \method and some baselines. 

%===============================================================================

\section{Limitations and Conclusion}
\label{sec:conclusion}
While promising, the \method approach has some limitations that motivate exciting directions for future work. 
% First, it is difficult to get high success rates during the hand poses synthesis phase, and thus our method has room to improve for more data-efficient sampling. 
% Second, we do not consider orientation when we evaluate pushing or pulling in the real world, drawing an incomplete picture of performance. 
% Third, we only study pushing and pulling as examples of nonprehensile manipulation, which does not exhaustively characterize all possible nonprehensile manipulation procedures. 
% Finally, it would be an interesting next step to make nonprehensile pushing or pulling truly closed-loop so it can react in real-time to unexpected disturbances such as object toppling. 
% First, an important direction for future work is to develop closed-loop pushing and pulling that can adapt in real time to environment interactions. Second, we only study pushing and pulling as examples of nonprehensile manipulation, which does not exhaustively characterize all possible nonprehensile manipulation procedures. To be specific, we aim to expand to a wider range of nonprehensile skills, including tilting or rolling, with dexterous hands, leveraging environmental interactions to augment dexterity. 
First, our study considers only pushing and pulling as examples of nonprehensile manipulation and thus does not exhaustively characterize all possible nonprehensile manipulation procedures; we aim to expand to a broader repertoire (e.g., tilting, rolling) that leverages environmental interactions with dexterous hands to enhance nonprehensile dexterity. Second, we aim to develop a closed-loop nonprehensile manipulation policy that adapts in real time to different object physical properties and environmental changes.
% We believe such capabilities will constitute a critical component of dexterous manipulation, extending its effectiveness beyond grasping.

To conclude, we propose \method, a dataset and method for nonprehensile object pushing and pulling using high-DOF robotic hands, such as the Allegro and LEAP Hands.  % Daniel: a bit tricky to decide whether we should capitalize H or not in 'hands' but I think if we consistently use H then it is OK. 
Our extensive real-world results show that \method enables diverse and effective pre-contact hand poses for different combinations of objects and pushing directions. We also demonstrate its usage for multi-step planning. 
We hope this inspires future work on dexterous nonprehensile manipulation.

\section{Acknowledgments}
We thank Wenhao Liu and Hanyang Zhou for their help with 3D printing objects. We are also grateful to everyone in the SLURM Lab and the RESL Lab at USC for their valuable discussions and feedback on problem formulation, experiments, and writing.

%===============================================================================

%===============================================================================

%% % The acknowledgments are automatically included only in the final and preprint versions of the paper.
%% \acknowledgments{If a paper is accepted, the final camera-ready version will (and probably should) include acknowledgments. All acknowledgments go at the end of the paper, including thanks to reviewers who gave useful comments, to colleagues who contributed to the ideas, and to funding agencies and corporate sponsors that provided financial support.}

%===============================================================================

% no \bibliographystyle is required, since the corl style is automatically used.
% \bibliographystyle{IEEEtranS}
% \bibliography{example}
\renewcommand{\bibfont}{\footnotesize}  
\printbibliography

\clearpage
\section{Additional Details of \method}
\label{app:method}

\subsection{Dataset Generation and Statistical Analysis}

\begin{table*}[hbp]
  \setlength\tabcolsep{4.6pt}
  \centering
    \footnotesize
    \begin{tabular}{lccc}
    \toprule
    Embodiment Part & \multicolumn{1}{c}{Finger Tip} & \multicolumn{1}{c}{Finger Link} & \multicolumn{1}{c}{Palm} \\ 
    \midrule
    Link No. & tip\_1, tip\_2, tip\_3, tip\_4 & 1,2,3,5,6,7,9,10,11,14,15 & palm\_link \\ 
    \midrule
    Number of Contact Candidates / each & 96  & 16 & 128 \\ 
    \bottomrule
    \end{tabular}
  \caption{Number of contact candidates on different parts of the Allegro Hand. We specify potential contacts all over the hand to encourage whole-hand (especially palm) nonprehensile manipulation on the object. See Table~\ref{tab:contact_candidates_leap} for that of the LEAP Hand.
  % \daniel{Maybe see if we can avoid ``tip\_3'' text, it looks a little awkward -- also why are they numbered that way?} \yunshuang{it indicates the finger tip on top of the link (tip\_3 is on top of line\_3)}
  }
  \vspace*{-5pt}
  \label{tab:contact_candidates_allegro}
\end{table*}  

\begin{table*}[htbp]
  \setlength\tabcolsep{4.6pt}
  \centering
    \footnotesize
    \begin{tabular}{lccccc}
    \toprule
    Embodiment Part & \multicolumn{1}{c}{Finger Tip} & \multicolumn{1}{c}{mcp\_joint \& dip} &
    \multicolumn{1}{c}{dip} &
    \multicolumn{1}{c}{thumb\_pip \& thumb\_dip} &
    \multicolumn{1}{c}{Palm} \\ 
    \midrule
    Link No. & tip\_1, tip\_2, tip\_3, thumb\_tip & 1,2,3 & 1,2,3 & -- & palm\_link \\ 
    \midrule
    Number of Contact Candidates / each & 24  & 16 & 4 & 16 & 128 \\ 
    \bottomrule
    \end{tabular}
  \caption{Number of contact candidates on different parts of the LEAP hand. See Table~\ref{tab:contact_candidates_allegro} for the Allegro Hand.
  % \daniel{Maybe see if we can avoid ``tip\_3'' text, it looks a little awkward -- also why are they numbered that way?} \yunshuang{it indicates the finger tip on top of the link (tip\_3 is on top of line\_3)}
  }
  \vspace*{-5pt}
  \label{tab:contact_candidates_leap}
\end{table*}  

\begin{figure}[h]
\centering \includegraphics[width=\linewidth]{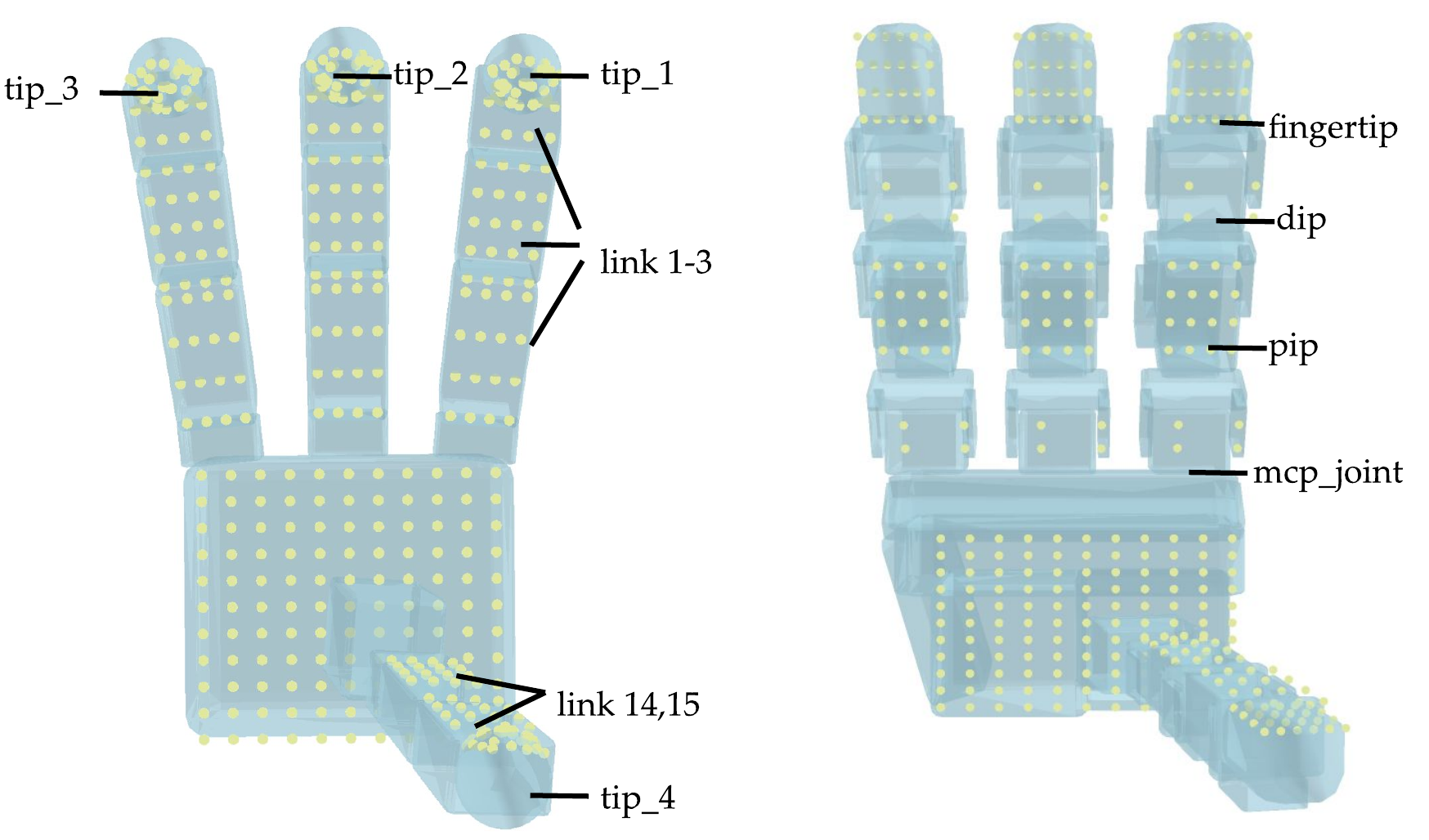}
  \caption{
    Contact candidates on the Allegro Hand and LEAP Hand. Refer to Table~\ref{tab:contact_candidates_allegro} and Table~\ref{tab:contact_candidates_leap} for the number of contacts on each link.
  }
  \label{fig:contact_candidates}
\end{figure}

\begin{figure}[h] 
  \centering
  \includegraphics[width=\linewidth]{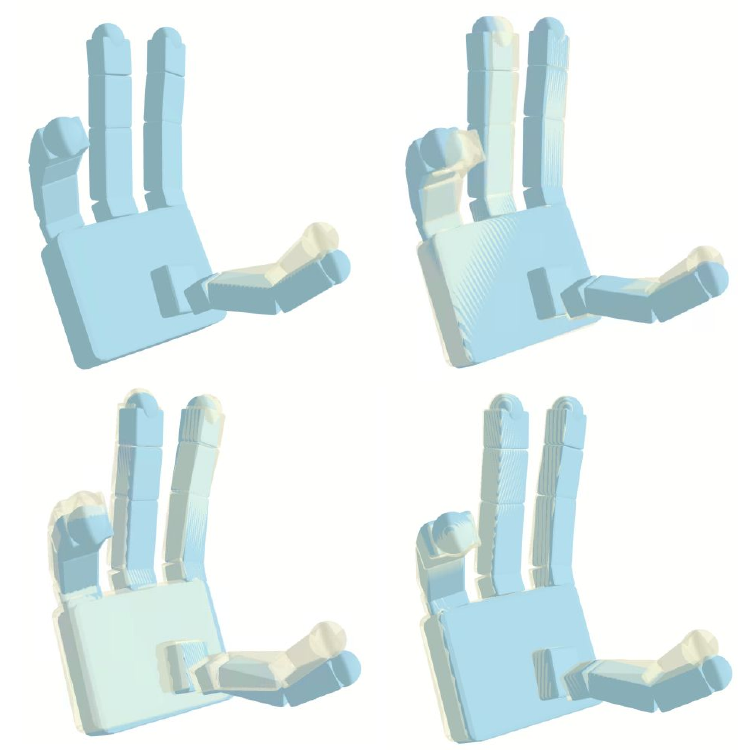}
  \caption{
  A visualization of an example of augmentations. \emph{Lightyellow} indicates the hand pose with the perturbation, and \emph{lightblue} is the original one.}
  \label{fig:noise_hand_pose}
\end{figure}

During dataset generation, we specify the contact candidates according to Figure~\ref{fig:contact_candidates} and Table~\ref{tab:contact_candidates_allegro}\&\ref{tab:contact_candidates_leap}, and we set the weight parameters (from Eq.~\ref{eq:energy}) according to values listed in Table~\ref{tab:weight_parameters}. For the optimization we discussed in Sec.~\ref{ssec:hand_poses}, the detailed hyperparameters are in Table~\ref{tab:optimization_hyperparameters}.

In the original hand pose generation procedure, we mainly consider the object geometry and encourage contact between selected contact candidates all over the hand and the object surface. However, it is crucial to test pushing to validate the quality of the nonprehensile hand poses. Initially, we obtain a low success rate of all generated hand poses, so we augment each successful hand pose 10 times. These perturbations involve small changes in rotation (max 2.5 deg), translation (max \SI{0.005}{\meter}) and joint pose (0.05 rad) using a Halton sequence. Figure~\ref{fig:noise_hand_pose} shows an example of a random original hand pose (lightblue color) and 4 different perturbed hand poses (lightyellow color). 
%In this way, we obtain the final dataset that has 58\% of them are successful and feed the successful ones for training the diffusion model, increasing data efficiency and scaling up the dataset.
By doing so, we get a large dataset of only successful hand poses, which we use for training the diffusion model.

\begin{figure*}[t]
    \centering
    \includegraphics[width=0.6\linewidth]{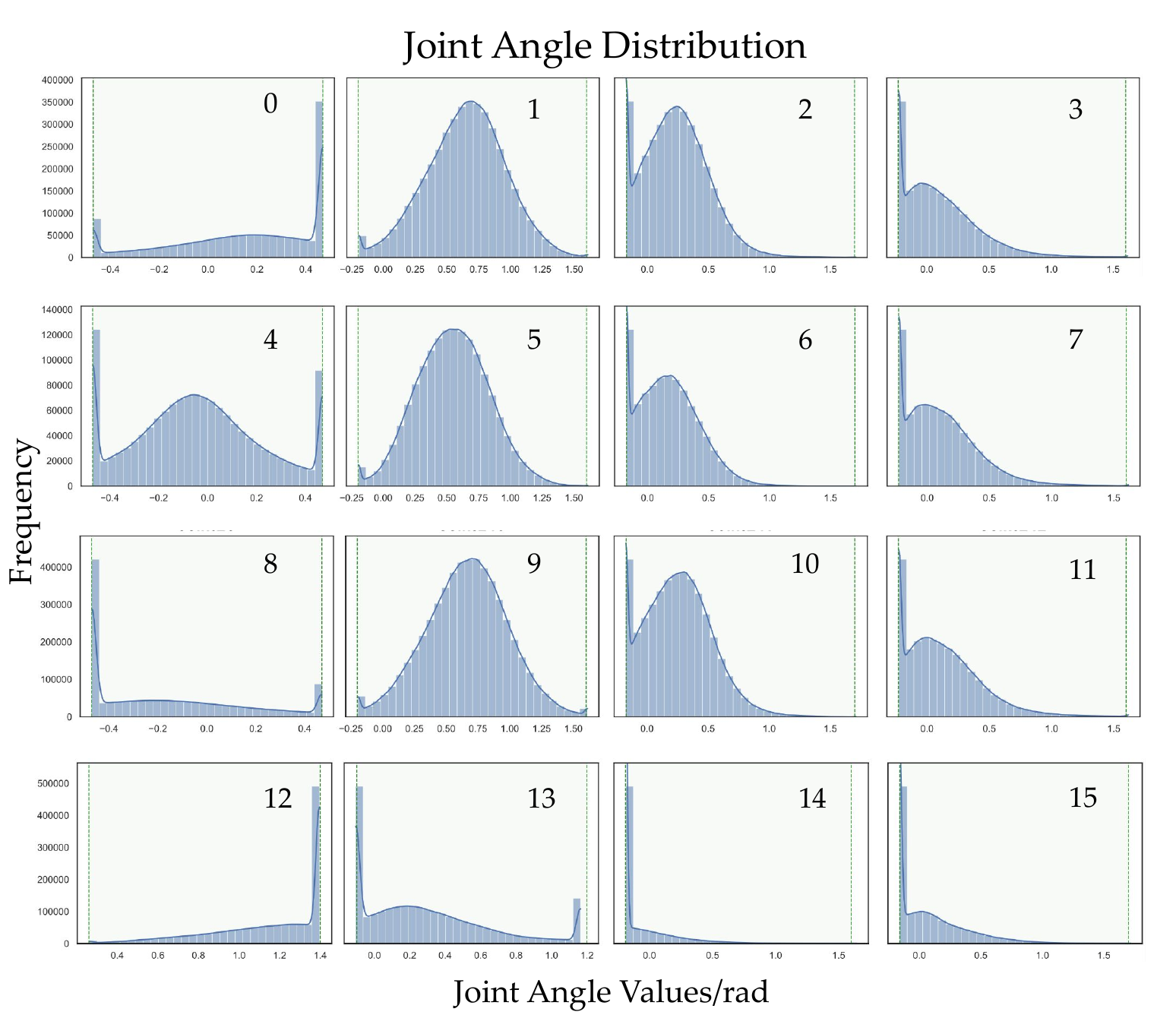}
    \caption{Visualization of the distribution of joint angle values in our proposed dataset, demonstrating the diversity of our generated hand poses. The number on the top right corner of each subfigure indicates the joint index. The \emph{green dashed lines} on the edge of x-axis indicate the lower/upper bounds of each joint angle values.}
    \label{fig:joint_angle_distribution}
    \vspace{-5pt}
\end{figure*}

\begin{figure*}[t]
    \centering
    \includegraphics[width=0.7\linewidth]{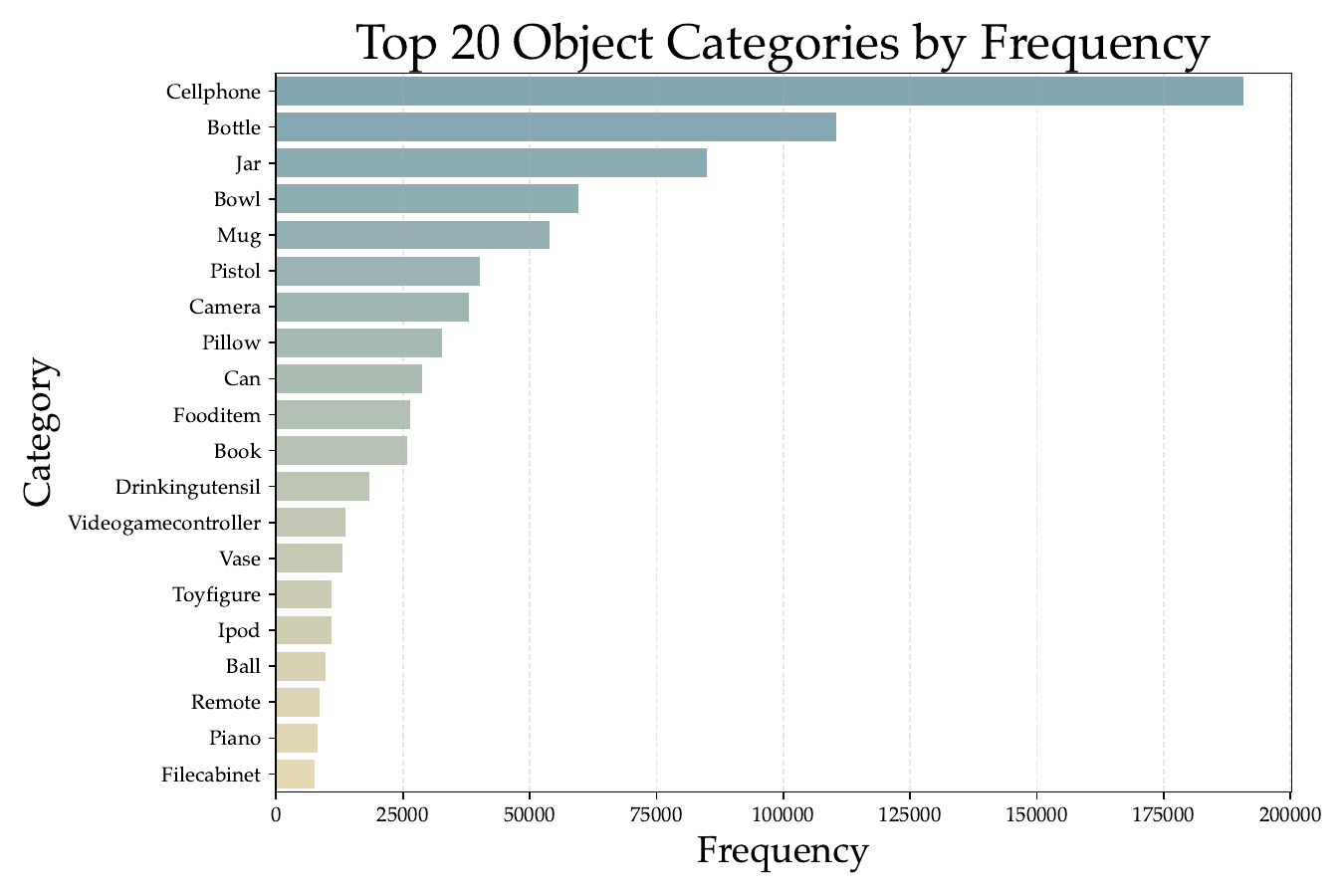}
    \caption{Visualization of the top 20 objects in terms of pushing hand poses frequency in our proposed dataset.}
    \label{fig:object_category}
\end{figure*}

Figure~\ref{fig:joint_angle_distribution} shows the distribution of joint angle values across our dataset. Most joints span the full range between their lower and upper bounds, and tend to have one or several modes. 
Those modes may lead to ``general'' stable hand poses for pushing motions. Other joint values may vary depending on particular object geometries.  
Figure~\ref{fig:object_category} shows a breakdown of object categories and the frequency of the top 20 objects in our dataset.

\begin{table}[H]
\centering
\begin{tabular}{ll}
\toprule
Parameter & Value \\
\midrule
$w_{\rm fc}$ & 0.5 \\
$w_{\rm dis}$ & 500 \\
$w_{\rm pen}$ & 300.0 \\
$w_{\rm spen}$ & 100.0 \\
$w_{\rm joints}$ & 1.0 \\
$w_{\rm ff}$ & 3.0 \\
$w_{\rm fp}$ & 0.0 \\
$w_{\rm tpen}$ & 100.0 \\
$w_{\rm direction}$ & 200.0 \\
$w_{\rm kinematics}$ & 100.0 \\
\bottomrule
\end{tabular}
\vspace{10pt}
\caption{Weight parameters.}
\label{tab:weight_parameters}
\end{table}

\begin{table}[H]
\centering
\begin{tabular}{ll}
\toprule
Parameter & Value \\
\midrule
Switch Possibility & 0.5 \\
$\mu$ & 0.98 \\
Step Size & 0.005 \\
Stepsize Period & 50 \\
Starting Temp. & 18 \\
Annealing Period & 30 \\
Temp. Decay & 0.95 \\
\bottomrule
\end{tabular}
\vspace{10pt}
\caption{Optimization hyperparameters.}
\label{tab:optimization_hyperparameters}
\end{table}

\subsection{Training Details}
\label{ssec:app_training_details}

\begin{figure*}[t]
    \centering
    \includegraphics[width=0.75\linewidth]{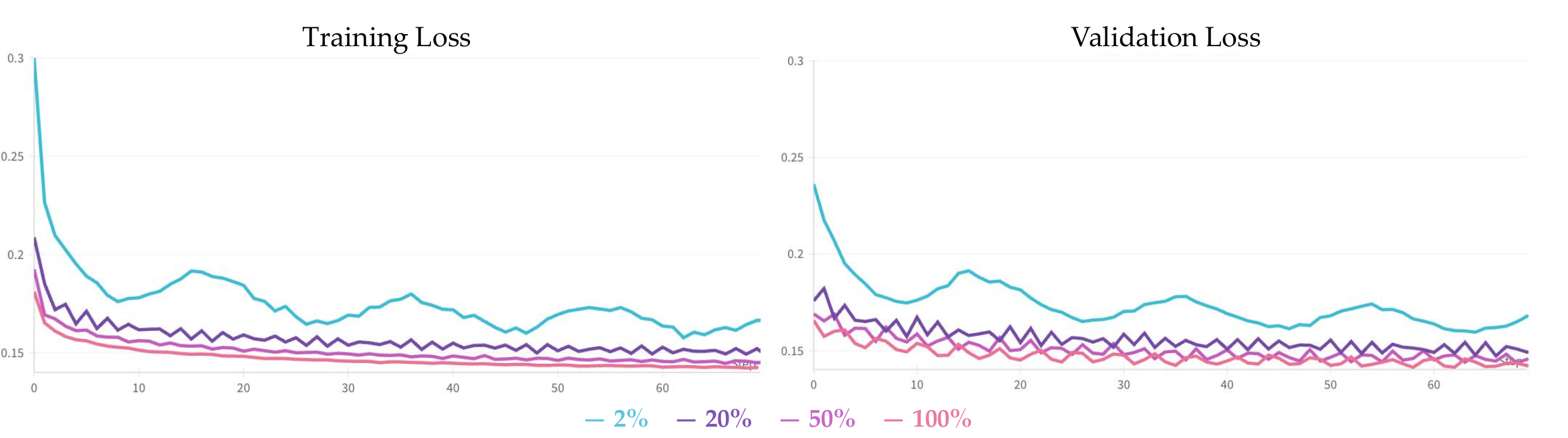}
    \caption{Training curves on different scales of the dataset. See Sec.\ref{ssec:simulation_experiments} for more discussion.}
    \label{fig:training_curves}
\end{figure*}

We train our model with one NVIDIA 4090 GPU on a desktop. Detailed training and model parameters are shown in Table~\ref{tab:training_parameters}. We also show the training curves with training loss and validation loss on different scales of the dataset in Figure~\ref{fig:training_curves}, which is relevant to our experiments in Sec.~\ref{ssec:simulation_experiments}.

\begin{table*}[htbp]
  \centering
  \begin{tabular}{lll}
    \toprule
    \textbf{Component} & \textbf{Parameter} & \textbf{Default / value} \\
    \midrule
    \multirow{2}{*}{\texttt{Data Config}}
        & \texttt{observation\_dim}        & 4096 \\
        & \texttt{pushingpose\_dim}    & 25  \\
    \midrule
    \multirow{3}{*}{\texttt{Model Config}}
        & \texttt{name}            & ConditionalUnet1D \\
        & \texttt{input\_dim}      & 25 \\
        & \texttt{global\_cond\_dim} & 4096 \\
    \midrule
    \multirow{4}{*}{\texttt{DDPM Scheduler}}
        & \texttt{beta\_schedule}         & squaredcos\_cap\_v2 \\
        & \texttt{clip\_sample}          & \texttt{True} \\
        & \texttt{num\_diffusion\_timesteps} & 100 \\
        & \texttt{prediction\_type}      & epsilon \\
    \midrule
    \multirow{4}{*}{\texttt{Training Config}}
        & \texttt{batch\_size}    & 16 \\
        & \texttt{n\_epochs}      & 200 \\
        & \texttt{print\_freq}    & 10 \\
        & \texttt{snapshot\_freq} & 25 \\
        % & \texttt{pred\_horizon}  & 1 \\
    \midrule
    \multirow{7}{*}{\texttt{Optim Config}}
        & \texttt{optimizer}   & Adam \\
        & \texttt{lr}          & \(1\times10^{-4}\) \\
        & \texttt{weight\_decay}& \(1\times10^{-6}\) \\
        & \texttt{beta1}       & 0.9 \\
        & \texttt{amsgrad}     & \texttt{False} \\
        & \texttt{eps}         & \(1\times10^{-8}\) \\
        & \texttt{grad\_clip}  & 1.0 \\
    \midrule
    \multirow{2}{*}{\texttt{lr Scheduler}}
        & \texttt{name}              & cosine \\
        & \texttt{num\_warmup\_steps} & 500 \\
    \midrule
    \multirow{1}{*}{\texttt{EMAModel}}
        & \texttt{power} & 0.75 \\
    \bottomrule
  \end{tabular}
  \caption{Configuration and training hyperparameters of the diffusion model.}
    \label{tab:training_parameters}
\end{table*}

% \subsection{Parameter Tuning}
% \label{ssec:app_parameters}

% \yunshuang{However it's tricky to tune the parameter do we mention it here or somewhere? but we did't really tune it carefully}

% \daniel{moved to appendix}

\section{Additional Details of Experiments}
\label{app:experiments}

\subsection{Experiment Details}

\begin{figure}[htbp]
    \centering
    \includegraphics[width=0.95\linewidth]{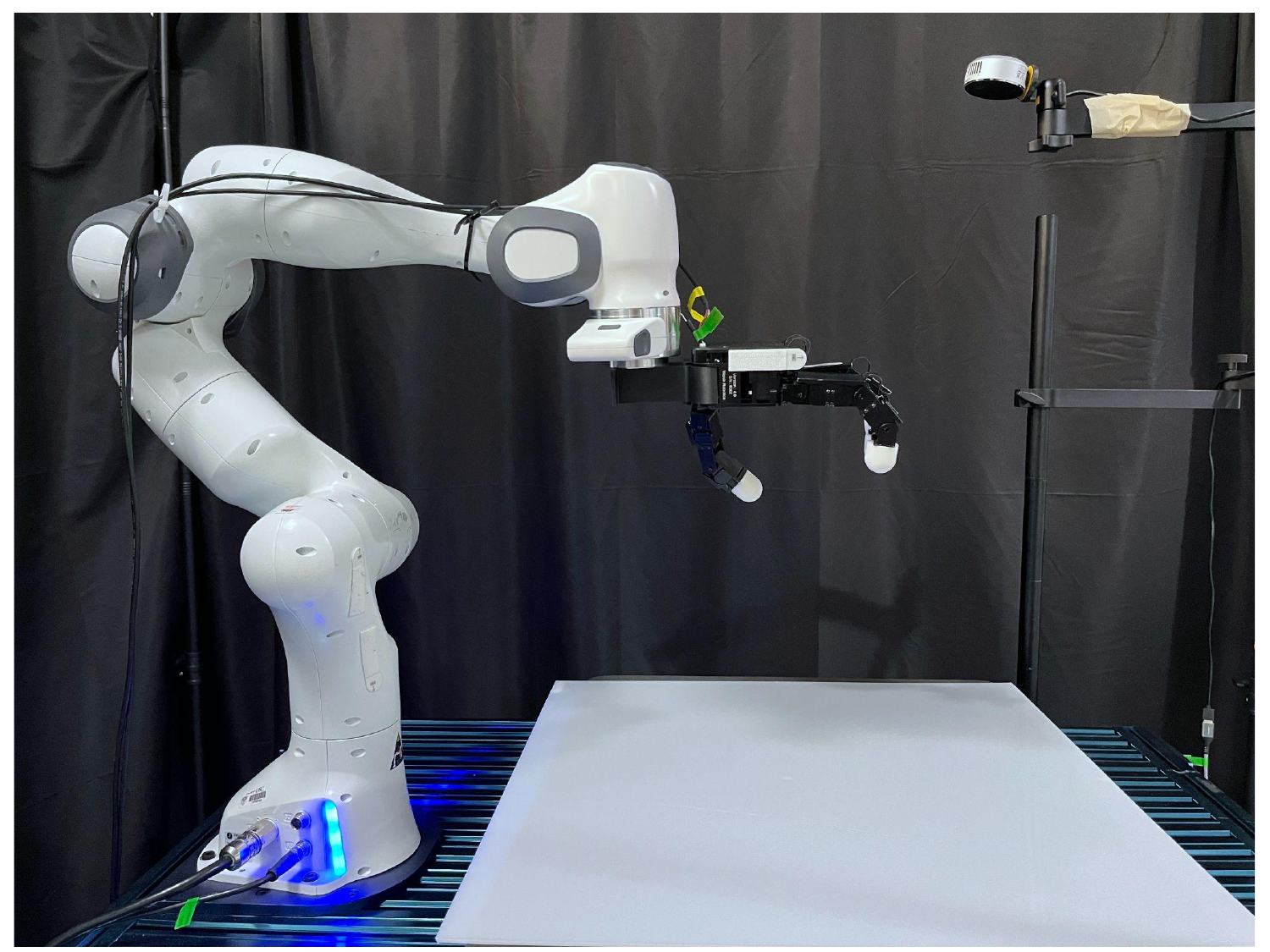}
    \caption{Our physical experiment setup including a Frank Panda robot with an attached Allegro Hand. The camera is only used for high-level path planning.}
    \label{fig:hardware_setup}
\end{figure}

Our physical experiment setup consists of a Franka Panda manipulator equipped with an Allegro Hand, as shown in Figure~\ref{fig:hardware_setup}. We also place an L515 RealSense camera above the table, which is \emph{only} used for path planning in multi-step planning experiments in Sec.~\ref{ssec:real_results} and Sec.~\ref{ssec:multi_step}. 
The surface we use for all experiments is a commercially available product purchased from Amazon (\href{https://www.amazon.com/dp/B0D2LC3YBF/ref=sspa_dk_detail_2?pd_rd_i=B0D2LCSHXR&pd_rd_w[…]956927c1&s=industrial&sp_csd=d2lkZ2V0TmFtZT1zcF9kZXRhaWw&th=1}{product\_link}). Since our focus is on nonprehensile hand pose generation, we assume that the surface's friction properties are sufficient to support pushing interactions. We leave a more detailed investigation of how physical properties influence dexterous nonprehensile manipulation as future work.

\begin{figure*}[t]
    \centering    
    \includegraphics[width=0.75\textwidth]{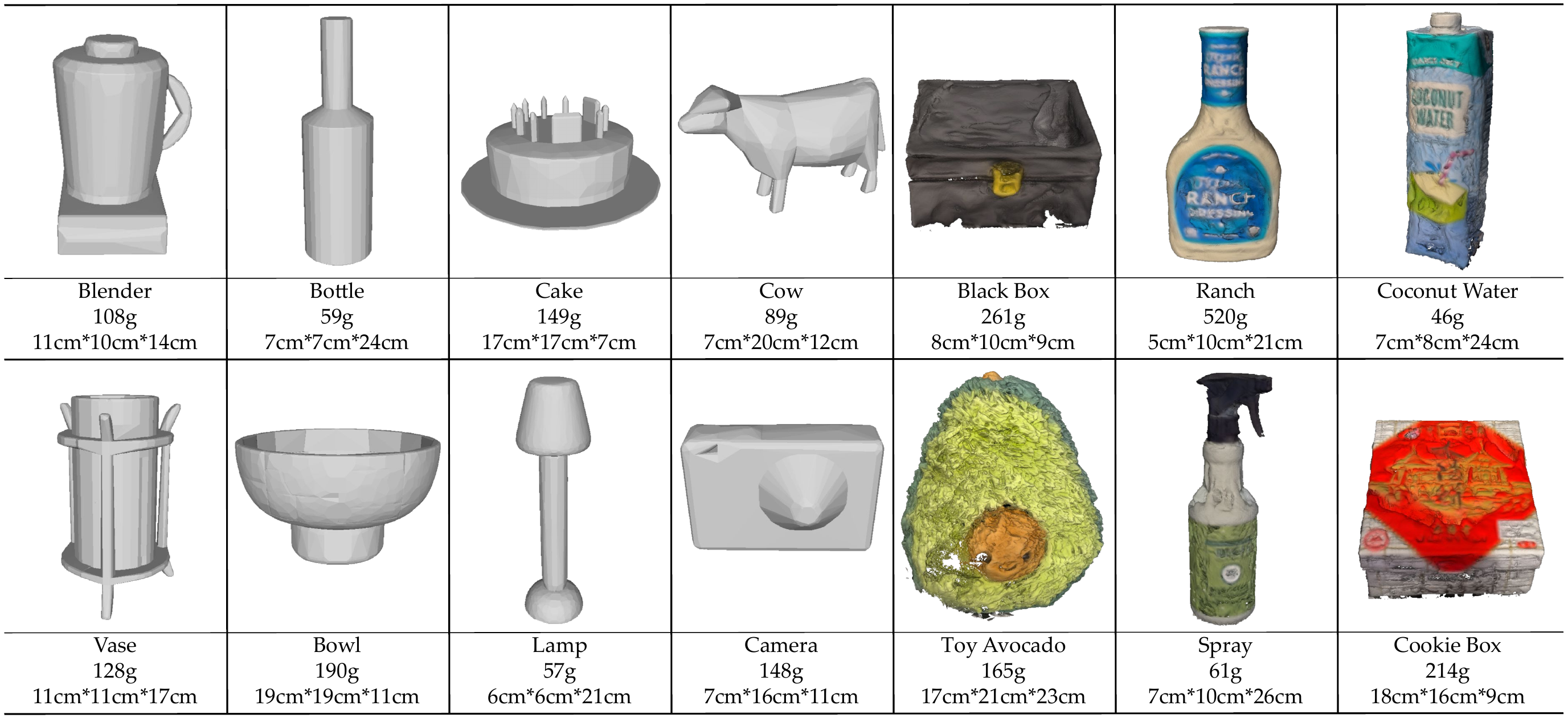}
    \caption{3D meshes, mass and physical dimensions of all objects tested in real-world experiments. Dimensions are listed as (x, y, z).}
    \label{fig:object_mesh}
\end{figure*}

We select 8 3D-printed objects and 6 real-world objects, covering flat, volumetric, and tall objects, as shown in Figure~\ref{fig:object_mesh}. Each object presents unique challenges for pushing. For example, when the robot hand approaches flat objects (e.g., Cake, Cookie Box) it may risk colliding with the table. In addition, tall objects (e.g., Lamp, Spray) frequently topple during pushing due to a high center of mass. 
While our method also suffers from these failure modes (particularly object toppling), it outperforms baselines, which topple objects more frequently. This motivates our case study on using a fixed hand pose to push objects taller than \SI{20}{\centi\meter}. While fixed hand poses can reliably work for objects with simple geometries, they frequently fail on these taller objects. As discussed in Sec.~\ref{ssec:real_results}, our results highlight the need for hand poses that provide more stable object support for transporting. 

We list the number of successful trials out of 5 for each method and direction in Table~\ref{tab:trials}. A blank entry (-) indicates that the robot could not execute the motion due to kinematic infeasibility. While \method has marginally more infeasible trials than the baselines, this is expected because  \method generates diverse hand orientations beyond top-down poses.
All methods execute pushes for \SI{20}{\centi\meter}, which is relatively long within the robot's workspace, and this can be infeasible for many hand poses. 
In contrast, the Pre-Trained Grasp Pose baseline tends to result in consistently top-down hand poses, which are generally easier to execute due to reachability and kinematic constraints. 
Despite counting all kinematically infeasible trials as failures, \method outperforms the baseline methods, demonstrating its robustness on pushing or pulling tasks.

\begin{table*}[htbp]
    \scriptsize
  \centering  
  \begin{tabular}{l *{5}{ccc}}
    \toprule
    %— Top header: Groups spanning three sub-columns each
    & \multicolumn{3}{c}{\textbf{\method}}
    & \multicolumn{3}{c}{\textbf{\method w/o Ranking}}
    & \multicolumn{3}{c}{\textbf{Nearest Neighbor}}
    & \multicolumn{3}{c}{\textbf{Pre-Trained Grasp Pose}} \\
    %— Sub-header: three rows per group
    \cmidrule(lr){2-4} \cmidrule(lr){5-7} \cmidrule(lr){8-10} \cmidrule(lr){11-13} 
    & Dir.1 & Dir.2 & Dir.3
    & Dir.1 & Dir.2 & Dir.3
    & Dir.1 & Dir.2 & Dir.3
    & Dir.1 & Dir.2 & Dir.3\\
    \midrule
    Blender & 5/5 & 4/5 & 4/5 & 3/5 & 3/5 & 5/5 & 2/5 & 2/5 & 2/5 & 1/5 & 1/5 & 1/5 \\
    Vase    & 5/5 & 3/5 & 4/5 & 2/5 & 4/5 & 4/5 & 4/5 & 4/5 & 3/5 & 2/5 & 3/5 & 2/5 \\
    Bottle  & 4/5 & 4/5 & 5/5 & 3/5 & 3/5 & 3/5 & 0/5 & 4/5 & 3/5 & 3/5 & 2/5 & 2/5 \\
    Bowl    & 4/5 & 1/5 & -   & 4/5 & 1/5 & -   & 2/5 & 2/5 & 1/5 & 3/5 & 2/5 & 2/5 \\
    Cake      & 4/5 & 3/5 & 4/5 & 4/5 & 4/5 & 3/5 & 3/5 & 1/5 & 1/5 & 1/5 & 0/5 & 1/5 \\
    Lamp      & 1/5 & 1/5 & 1/5 & 2/5 & 2/5 & 2/5 & 1/5 & 0/5 & 0/5 & 0/5 & 1/5 & 1/5 \\
    Cow       & 5/5 & 3/5 & 3/5 & 3/5 & 2/5 & 3/5 & 1/5 & 1/5 & 1/5 & 0/5 & 3/5 & 2/5 \\
    Camera    & 2/5 & 2/5 & 4/5 & 2/5 & 3/5 & 3/5 & 1/5 & 1/5 & 3/5 & 1/5 & 4/5 & 2/5 \\
    \midrule
    3D Avg./ \% & 67.5 & 52.5 & 62.5 & 57.5 & 55.0 & 57.5 & 35.0 & 37.5 & 35.0 & 27.5 & 40.0 & 32.5 \\
    \midrule
    Black Box  & 4/5 & 4/5 & 3/5 & 3/5 & 1/5 & 2/5 & 1/5 & 1/5 & 2/5 & 3/5 & 3/5 & 2/5 \\
    Toy Avocado  & 4/5 & -   & 1/5 & 3/5 & - & 2/5 & - & - & 1/5 & 3/5 & 0/5 & 4/5 \\
    Ranch     & 3/5 & 2/5 & 3/5 & 4/5 & 1/5 & 2/5 & 3/5 & 1/5 & 4/5 & 1/5 & - & 2/5 \\
    Spray    & 3/5  & -   & 1/5 & 0/5 & - & 1/5 & 2/5 & - & 2/5 & 0/5 & 0/5 & 2/5 \\
    Coconut Water& 2/5 & 3/5 & 4/5 & 2/5 & 2/5 & 1/5 & 2/5 & 1/5 & 2/5 & 0/5 & 0/5 & 0/5 \\
    Cookie Box & -   & 5/5 & 3/5 & -   & 2/5 & 5/5 & -   & 3/5 & 2/5 & 2/5 & 2/5 & 1/5 \\
    \midrule
    DO Avg./ \% &
    53.3 & 40.0 & 50.0 & 40.0 & 20.0 & 43.3 & 30.0 & 16.7 & 30.0 & 26.7 & 20.0 & 43.3 \\
    \midrule
    All Avg./ \%& 61.4 & 47.1 & 57.1 &
    50.0 & 40.0 &
    51.4 &
    32.9 &
    28.6 &
    32.9 &
    27.1 &
    31.4 &
    37.1\\
    \bottomrule
  \end{tabular} 
  
  \caption{Detailed experiment results for each object and direction combination. ``3D Avg.'' refers to the average success rate over all 3D-printed objects, ``DO Avg.'' is that of daily objects and ``All Avg.'' is that of all 14 test objects. These results correspond to the bar charts in  Figure~\ref{fig:bar_plot_baselines}.}
  \label{tab:trials}
\end{table*}

\subsection{More Successful Rollouts}

\begin{figure*}[t]
    \centering
    \includegraphics[width=0.75\textwidth]{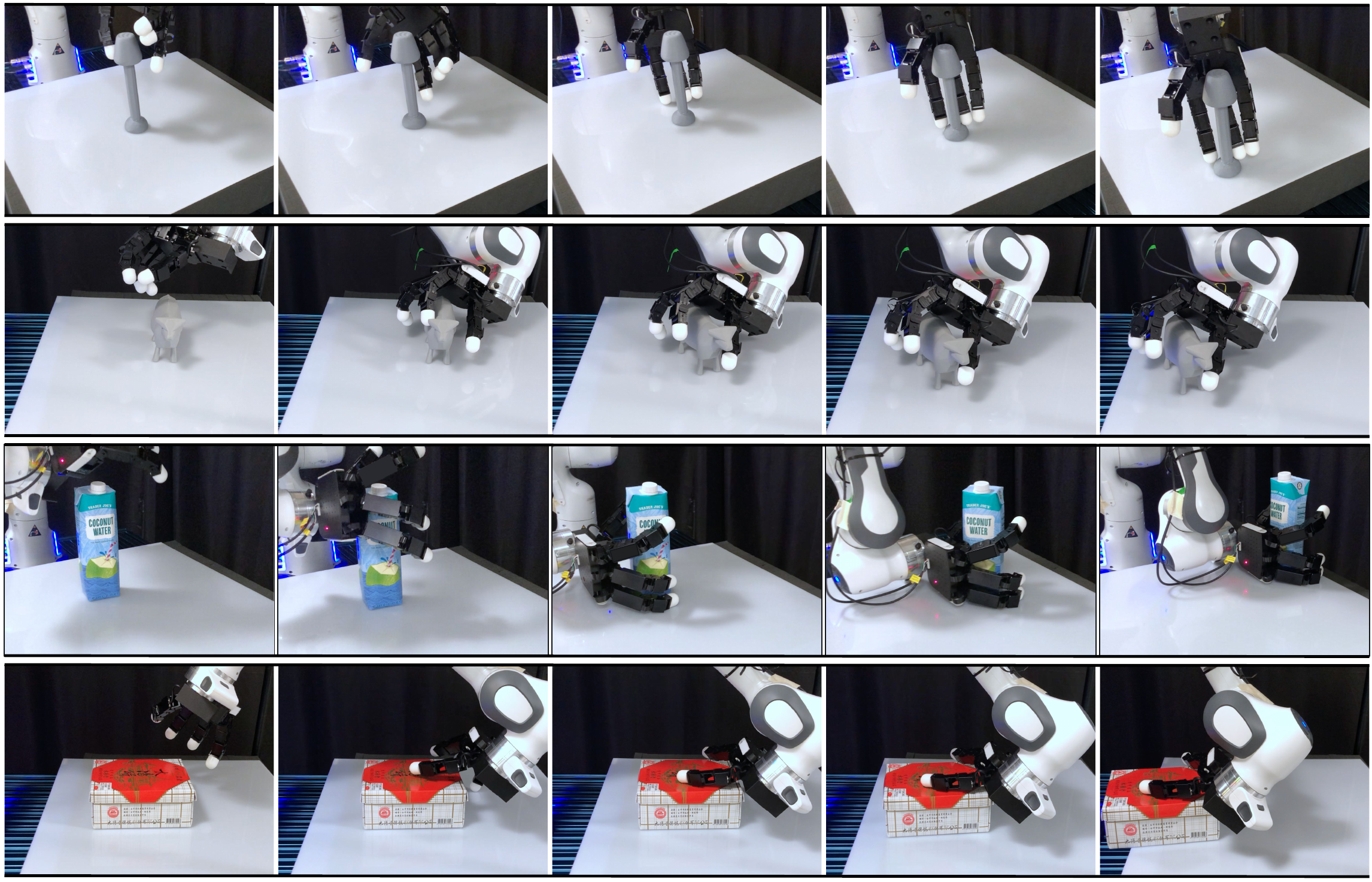}
    \caption{Successful rollouts of \method, one per row.}
    \label{fig:success_rollouts}
\end{figure*}

We provide additional example visualizations of successful rollouts of \method in Figure~\ref{fig:success_rollouts}. For videos, please refer to our website: \href{https://geodex2p.github.io/}{geodex2p.github.io}. 

\subsection{Results and Analysis of Baseline Methods}

\begin{figure*}[htbp]
    \centering    
    \includegraphics[width=0.75\textwidth]{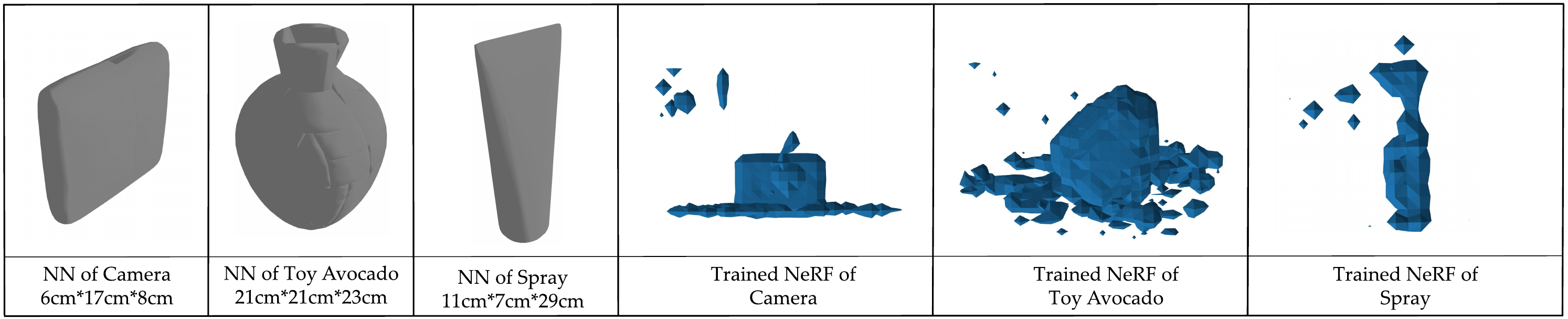}
    \caption{Nearest Neighbor retrieval results of three test objects (left three columns). The retrieved objects share geometric similarities with the query object: a box-shaped match for the camera, a spherical-shaped vase for the toy avocado, and a tall, slender shape for the spray. The pre-trained grasp pose baseline takes in NeRF as input(right three columns). We trained the NeRF for our objects following their instructions and pipelines~\cite{lum2024gripmultifingergraspevaluation}.}
    \label{fig:baseline_figure}
\end{figure*}

We visualize 3 examples of the nearest neighbor (NN) retrieval results and the trained NeRF representation in Figure~\ref{fig:baseline_figure}. The retrieved NN objects are similar in shape and scale of the query object (left 3 columns in Figure~\ref{fig:baseline_figure}). However, their coarse geometry granularity is insufficient to generate robust hand poses. For example, with the Toy Avocado, our method selects a hand pose that pushes from the bottom to avoid sliding or toppling. 
In contrast, the NN method retrieves a vase-like object, where pushes from the middle make more sense. The irregular geometric shape at the bottom of the vase-like object could potentially cause more collisions and may increase the difficulty of solving the kinematics. 
The right 3 columns in Figure~\ref{fig:baseline_figure} visualize the NeRF input to the Pre-Trained Grasp Pose method, since we use their pre-trained model taking in NeRF representations. Though a common failure mode of the pre‑trained grasp pose is that the object slips from the hand because the palm is oriented at an improper angle, we observe notable visual noise in the NeRF representation, which may also deteriorate performance of this baseline. For more discussions of baseline performance, see Sec.~\ref {ssec:real_results}. 

% \begin{table*}[h]
%   \centering  \label{tab:grasp_success}
%   \begin{tabular}{lcccccc}
%     \toprule
%     Object & Bottle & Lamp & Coconut Water & Ranch & Spray & Toy Avocado \\ \midrule
%     Successes / Trials & 9/10 & 4/10 & 0/10 & 0/10 & 0/10 & 5/10 \\ \bottomrule
%   \end{tabular}
%   \caption{Grasp‑success statistics for each object (successes out of 10 trials).}
%   \vspace*{-14pt}
% \end{table*}

\subsection{Multi-step Planning}
\label{ssec:multi_step} 

% \begin{figure}[H]
% \centering
% \includegraphics[width=\linewidth]{figures/multistage_merged.pdf}
% \caption{
% Path planning using RRT* for multi-step planning. The first column shows the visualization of path planning results. The second and third columns show two consecutive hand poses for pushing the object along the path. The first example is the same as the one shown in Fig.~\ref{fig:results_multistep}.
% }
% \label{fig:multistep_path}
% \vspace*{-5pt}
% \end{figure}

% \begin{figure*}[h]
% \vspace*{-10pt}
% \centering
% \begin{minipage}{0.42\textwidth}
%   \centering
%   \includegraphics[width=\textwidth]{figures/multistage_path_v02.pdf}
%   \caption{
%     Contact candidates on the Allegro hand. Refer to Table~\ref{tab:contact_candidates} for the detailed number of contacts on each link.
%   }
%   \label{fig:contact_candidates}
% \end{minipage}
% \hfill
% \begin{minipage}{0.49\textwidth}
%   \centering
%   \includegraphics[width=\textwidth]{figures/multistage_v02.pdf}
%   \caption{
%   A visualization of an example of augmentations. \emph{Lightyellow} indicates the hand pose with the perturbation, and \emph{lightblue} is the original one.}
%   \label{fig:noise_hand_pose}
% \end{minipage}
% \vspace*{-10pt}
% \end{figure*}

% \yunshuang{The pictures are taken with the l515 realsense camera in Figure~\ref{fig:hardware_setup}. We first did the calibration xxx then calculate the path, how to define start and goal, how to detect the obstacle, how to set the obstacle radius in rrt*}

Here, we provide more information and context on top of the \emph{Multi-step Planning} section in Sec.~\ref{ssec:real_results}.
These experiments explore the potential for \method's hand poses to support long-horizon planning.  
As shown in Figure~\ref{fig:results_multistep}, an Intel RealSense L515 camera captures a top-down view of the scene (see Figure~\ref{fig:results_multistep}). A toy placed in the scene serves as an obstacle. We extract its segmentation mask using Grounded SAM 2~\cite{liu2023grounding,ravi2024sam,ren2024grounded,kirillov2023segany,jiang2024trex2}, define the toy's position at its (estimated) center, and set a fixed \SI{20}{\centi\meter} radius for path planning. The start and goal positions are manually assigned. We use RRT* as a high-level planner to compute a collision-free path in the 2D image space. Through camera calibration, we convert the 2D waypoints into 3D coordinates in the robot frame. For each edge along the planned path, \method generates a corresponding hand pose, and the robot pushes the object towards the next waypoint.

We test with two episodes that cover more pushing directions. The key insight in these experiments is that hand poses should be considered and evaluated while considering the kinematics of the arm as the motion becomes more complex. In the second row of Figure~\ref{fig:results_multistep}, a similar hand pose is able to finish the two-step pushing tasks while avoiding the obstacle. However, the first row of Figure~\ref{fig:results_multistep} shows the need to change hand poses to better fit the object pose and the intended pushing direction. This motivates our use of motion planning and pose ranking to facilitate stable and smooth multi-step pushing motions. 

% \bibliography{supplement}

\end{document}